\documentclass{article}

\usepackage{microtype}
\usepackage{graphicx}
\usepackage{subfigure}
\usepackage{booktabs} %

\usepackage[accepted]{icml2024}

\usepackage{color,xcolor}
\usepackage{epsfig}
\usepackage{graphicx}

\usepackage{adjustbox}
\usepackage{array}
\usepackage{booktabs}
\usepackage{colortbl}
\usepackage{float,wrapfig}
\usepackage{hhline}
\usepackage{multirow}
\usepackage[font={small}]{caption}
\usepackage{natbib}

\usepackage{amsmath,amsfonts,amsthm,amssymb}
\usepackage{bm}
\usepackage{nicefrac}
\usepackage{microtype}
\usepackage{mathtools}

\usepackage{changepage}
\usepackage{extramarks}
\usepackage{fancyhdr}
\usepackage{lastpage}
\usepackage{setspace}
\usepackage{soul}
\usepackage{xspace}

\usepackage{url}

\usepackage{algorithmic}
\usepackage{algorithm}
\usepackage{todonotes} %
\usepackage{enumitem}
\usepackage{titlesec}
\usepackage{bbm}

\usepackage{amsmath,amsfonts,bm}

\def\eqref#1{equation~\ref{#1}}

\def\1{\bm{1}}

\def\eps{{\epsilon}}

\def\rb{{\textnormal{b}}}

\def\vb{{\bm{b}}}

\def\vt{{\bm{t}}}

\def\vx{{\bm{x}}}

\def\vz{{\bm{z}}}

\DeclareMathAlphabet{\mathsfit}{\encodingdefault}{\sfdefault}{m}{sl}
\SetMathAlphabet{\mathsfit}{bold}{\encodingdefault}{\sfdefault}{bx}{n}

\def\sqrtexplained#1{%
  \begingroup
    \sbox0{$#1$}
    \def\underbrace##1_##2{##1}
    \sbox2{$#1$}
    \dimen0=\wd0 \advance\dimen0-\wd2
    \mathrlap{\sqrt{\phantom{\displaystyle#1}\kern\dimen0 }}
    \hphantom{\sqrt{\vphantom{\displaystyle#1}}}
  \endgroup
  #1}

\def\Gcal{\mathcal{G}}
\def\Xcal{\mathcal{X}}
\def\Ccal{\mathcal{C}}
\def\Ncal{\mathcal{N}}
\def\Acal{\mathcal{A}}
\def\Scal{\mathcal{S}}

\renewcommand{\widehat}{\hat}

\newcommand{\myparagraph}[1]{\noindent\textbf{#1}}

\definecolor{rowblue}{RGB}{220,230,240}
\definecolor{myorchid}{RGB}{150,10,30}
\definecolor{myblue}{RGB}{10,30,250}
\definecolor{mygreen}{RGB}{10,120,10}

\definecolor{bubblegum}{rgb}{0.99, 0.76, 0.8}
\newcommand{\cm}[1]{{#1}}

\newcommand{\sect}[1]{Section~\ref{#1}}

\newcommand{\eqn}[1]{Equation~\ref{#1}}
\newcommand{\fig}[1]{Figure~\ref{#1}}

\usepackage{Definitions}

\usepackage{amsmath}
\usepackage{amssymb}
\usepackage{mathtools}
\usepackage{amsthm}

\usepackage[pagebackref=true,breaklinks=true,colorlinks,citecolor=gray]{hyperref}

\usepackage[capitalize,noabbrev]{cleveref}

\theoremstyle{plain}

\begin{document}

\twocolumn[
\icmltitle{Compositional Image Decomposition with Diffusion Models}

\icmlsetsymbol{equal}{*}

\begin{icmlauthorlist}
\icmlauthor{Jocelin Su}{yyy,equal}
\icmlauthor{Nan Liu}{comp,equal}
\icmlauthor{Yanbo Wang}{zzz,equal}
\icmlauthor{Joshua B. Tenenbaum}{yyy}
\icmlauthor{Yilun Du}{yyy}
\end{icmlauthorlist}

\icmlaffiliation{yyy}{MIT}
\icmlaffiliation{comp}{UIUC}
\icmlaffiliation{zzz}{TU Delft}

\icmlcorrespondingauthor{Jocelin Su}{jocelin@mit.edu}

\icmlkeywords{Machine Learning, ICML}

\vskip 0.3in
]

\printAffiliationsAndNotice{\icmlEqualContribution} %

\titlespacing\section{0pt}{3pt plus 1pt minus 1pt}{2pt plus 1pt minus 1pt}
\titlespacing\subsection{0pt}{2pt plus 1pt minus 1pt}{2pt plus 1pt minus 1pt}
\setlength{\parskip}{0.55em}

\setlength{\belowdisplayskip}{3pt}
\setlength{\abovedisplayskip}{3pt}
\setlength{\belowdisplayshortskip}{3pt}
\setlength{\abovedisplayshortskip}{3pt}

\def\methodname{Decomp Diffusion\xspace}
\def\model{Decomp Diffusion\xspace}

\begin{abstract}
Given an image of a natural scene, we are able to quickly decompose it into a set of components such as objects, lighting, shadows, and foreground. We can then envision a scene where we combine certain components with those from other images, for instance a set of objects from our bedroom and animals from a zoo under the lighting conditions of a forest, even if we have never encountered such a scene before. In this paper, we present a method to decompose an image into such compositional components. Our approach, \methodname, is an unsupervised method which, when given a single image, infers a set of different components in the image, each represented by a diffusion model. We demonstrate how components can capture different factors of the scene, ranging from global scene descriptors like shadows or facial expression to local scene descriptors like constituent objects. We further illustrate how inferred factors can be flexibly composed, even with factors inferred from other models, to generate a variety of scenes sharply different than those seen in training time. Code and visualizations are at \url{https://energy-based-model.github.io/decomp-diffusion}.
\end{abstract}

\section{Introduction}\label{sec:intro}
\begin{figure*}[t!]
\begin{center}
\includegraphics[width=\textwidth]{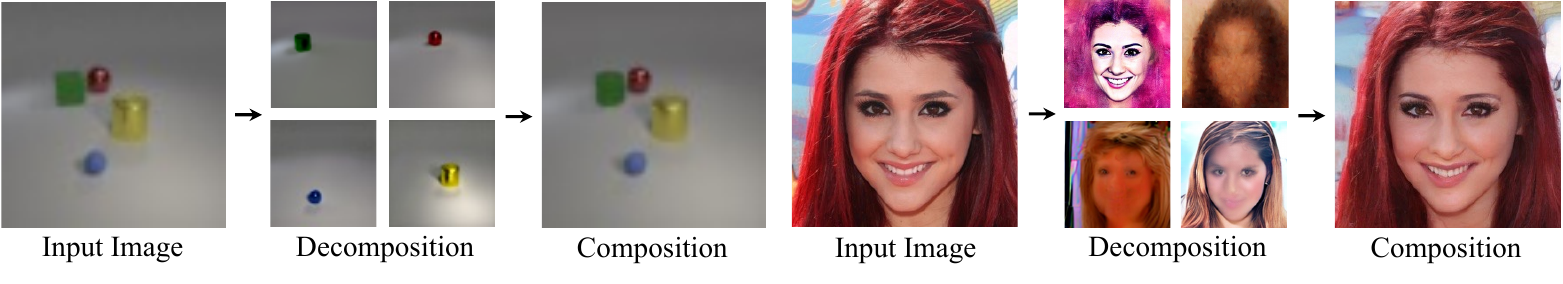}
\end{center}
\vspace{-10pt}
\caption{ \textbf{Image Decomposition with \model.} Our unsupervised method can decompose an input image into both local factors, such as objects (\textbf{Left}), and global factors (\textbf{Right}), such as facial features. Additionally, our approach can combine the deduced factors for image reconstruction.}
\label{fig:teaser}
\vspace{-15pt}
\end{figure*}

Humans have the remarkable ability to quickly learn new concepts, such as learning to use a new tool after observing just a few demonstrations \citep{Allen29302}. This skill relies on the ability to combine and reuse previously acquired concepts to accomplish a given task \citep{lake2017building}. This is particularly evident in natural language, where a limited set of words can be infinitely combined under grammatical rules to express various ideas and opinions \citep{chomsky1965}. In this work, we propose a method to discover  compositional concepts from images in an unsupervised manner, which may be flexibly combined both within and across different image modalities. 

Prior works on unsupervised compositional concept discovery may be divided into two separate categories. One line of approach focuses on discovering a set of global, holistic factors by representing data points in fixed factorized vector space~\citep{vedantam2017generative,higgins2017scan,singh2019finegan,peebles2020hessian}. Individual factors, such as facial expression or hair color, are represented as independent dimensions of the vector space, with recombination between concepts corresponding to recombination between underlying dimensions. However, since the vector space has a fixed dimensionality, multiple instances of a single factor, such as multiple different sources of lighting, may not be easily combined. Furthermore, as the vector space has a fixed underlying structure, individual factored vector spaces from different models trained on different datasets may not be combined, \eg, the lighting direction in one dataset with the foreground of an image from another.

An alternative approach decomposes a scene into a set of different underlying ``object" factors. Each individual factor represents a separate set of pixels in an image defined by a disjoint segmentation mask~\citep{burgess2019monet,locatello2020objectcentric,monnier2021unsupervised,engelcke2021genesis}. Composition between different factors then corresponds to composing their respective segmentation masks. However, this method struggles to model higher-level relationships between factors, as well as multiple global factors that collectively affect the same image.

Recently, COMET~\citep{du2021comet} proposes to instead decompose a scene into a set of factors represented as \textit{energy functions}. Composition between factors corresponds to solving for a minimal energy image subject to each energy function. Each individual energy function can represent global concepts such as facial expression or hair color as well as local concepts such as objects.
However, COMET is unstable to train  due to second-order gradients, and often generates blurry images. 

In this paper, we leverage the close connection between Energy-Based Models~\citep{lecun2006tutorial,du2019implicit} and diffusion models~\citep{sohl2015deep,ho2020denoising} and propose \model, an approach to decompose a scene into a set of factors, each represented as separate diffusion models. Composition between factors is achieved by sampling images from a composed diffusion distribution~\citep{liu2022compositional,du2023reduce}, as illustrated in Figure \ref{fig:teaser}. Similar to composition between energy functions, this composition operation allows individual factors to represent both global and local concepts and further enables the recombination of concepts across models and datasets.

However, unlike the underlying energy decomposition objective of COMET, \model may directly be trained through denoising, a stable and less expensive learning objective, and leads to higher resolution images. In summary, we contribute the following: First, we present \model, an approach using diffusion models to decompose scenes into a set of different compositional concepts which substantially outperforms prior work using explicit energy functions. Second, we show that \model is able to successfully decompose scenes into both global concepts as well as local concepts. Finally, we show that concepts discovered by \model generalize well, and are amenable to compositions across different modalities of data, as well as components discovered by other instances of \model.

\section{Unsupervised Decomposition of Images into Energy Functions}
\label{sec:background}

In this section, we introduce background information about COMET~\citep{du2021comet}, which our approach extends. COMET infers a set of latent factors from an input image, and uses each inferred latent to define a separate energy function over images. To generate an image that exhibits inferred concepts, COMET runs an optimization process over images on the sum of different energy functions.

In particular, given an image $\vx_i \in \mathbb{R}^D$, COMET uses a learned encoder $\text{Enc}_\cm{\phi}(\vx_i)$ to infer a set of $K$ different latents $\vz_k \in \mathbb{R}^M$, where each latent $\vz_k$ represents a different concept in an image. Both the image and latents are passed into an energy function $E_\theta(\vx_i, \vz_k): \mathbb{R}^D \times \mathbb{R}^M \rightarrow \mathbb{R}$, which maps these variables to a scalar energy value.

Given a set of different factors $\vz_k$, decoding these factors to an image corresponds to solving the optimization problem:
\begin{equation}
    \argmin_{\vx} \sum_k E_\theta(\vx; \vz_k).
\end{equation}
To solve this optimization problem, COMET runs an iterative gradient descent procedure from an image initialized from Gaussian noise. Factors inferred from either different images or even different models may likewise be decoded by optimizing the energy function corresponding to sum of energy function of each factor.

COMET is trained so that the $K$ different inferred factors $\vz_k$ from an input image $\vx_i$ define $K$ energy functions, so that the minimal energy state corresponds to the original image $\vx_i$:
\begin{equation}
      \label{eqn:loss}
\resizebox{0.75\hsize}{!}{$
      \mathcal{L}_{\text{MSE}}(\theta) = \left\| \argmin \limits_{\vx} \left( \sum_k E_{\theta}(\vx; \vz_k) \right) - \vx_i \right\|^2 $},
\end{equation}
where $\vz_k = \text{Enc}_\cm{\phi}(\vx_i)[k]$. The $\argmin$ of the sum of the energy functions is approximated by $N$ steps of gradient descent
\begin{equation}
    \vx_i^N = \vx_i^{N-1} - \gamma \nabla_{\vx} \sum_k E_{\theta} 
 (\vx_i^{N-1}; \text{Enc}_\cm{\phi}(\vx_i)[k]),
    \label{eqn:opt_approx}
\end{equation}
where $\gamma$ is the step size. Optimizing the training objective in \eqn{eqn:loss} 
 corresponds to back-propagating through this optimization objective. The resulting process is computationally expensive and unstable to train, as it requires computing second-order gradients.

\section{Compositional Image Decomposition with Diffusion Models}\label{sec:method}
\begin{figure*}[t!]
\begin{center}
\includegraphics[width=\textwidth]{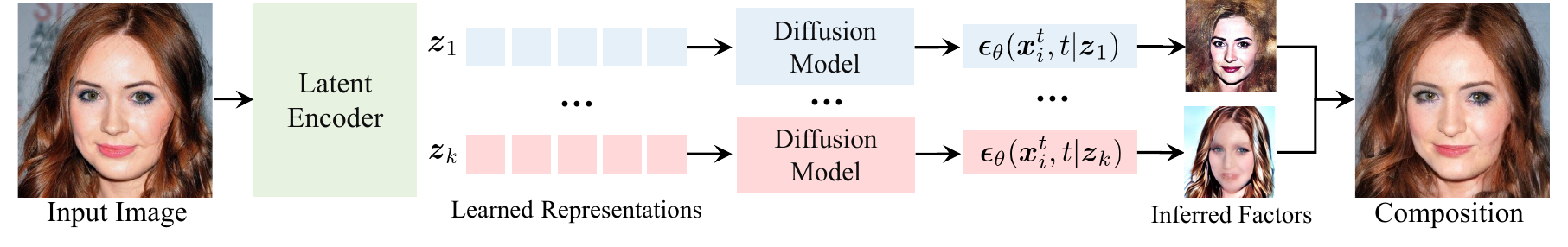}
\end{center}
\vspace{-10pt}
\caption{\small \textbf{Compositional Image Decomposition}. We learn to decompose each input image into a set of denoising functions $\{\epsilon_\theta(\vx_i^t, t, | \vz_k)\}$ representing $K$ factors, which can be composed to reconstruct the input.
}
\label{fig:model}
\vspace{-10pt}
\end{figure*}

Next, we discuss how to decompose images into a set of composable diffusion models. We first discuss how diffusion models may be seen as parameterizing energy functions in \sect{sect:diffusion_ebm}. Then in \sect{sect:decomp_diffusion}, we describe how we use this connection in \model to decompose images into a set of composable diffusion models.

\subsection{Denoising Networks as Energy Functions}
\label{sect:diffusion_ebm}

Denoising Diffusion Probabilistic Models (DDPMs)~\citep{sohl2015deep,ho2020denoising} are a class of generative models that facilitate generation of images $\vx_0$ by iteratively denoising an image initialized from Gaussian noise. Given a randomly sampled noise value $\epsilon \sim \mathcal{N}(0, 1)$, as well as a set of $t$ different noise levels $\epsilon^t = \sqrt{\beta_t} \epsilon$ added to a clean image $\vx_i$, a denoising model $\epsilon_\theta$ is trained to denoise the image at each noise level $t$:
\begin{equation}
 \mathcal{L}_{\text{MSE}}=\|\mathbf{\epsilon} - \epsilon_\theta(\sqrt{1-\beta_t}\vx_i +  \sqrt{\beta_t} \mathbf{\epsilon}, t)\|_2^2.
\end{equation}
In particular, the denoising model learns to estimate a gradient field of natural images, describing the direction that noisy images $\vx^t$ with noise level $t$ should be refined toward to become natural images~\citep{ho2020denoising}. As discussed in both ~\citep{liu2022compositional,du2023reduce}, this gradient field also corresponds to the gradient field of an energy function 
\begin{equation}
    \epsilon_\theta(\vx^t, t) = \nabla_{\vx} E_\theta(\vx)
    \label{eqn:score_energy}
\end{equation}
that represents the relative log-likelihood of a datapoint.

To generate an image from the diffusion model, a sample $\vx^T$ at noise level $T$ is initialized from Gaussian noise $\mathcal{N}(0, 1)$ and then iteratively denoised through
\begin{equation}
    \vx^{t-1}=\vx^{t}-\gamma\epsilon_\theta(\vx^t,t) + \xi, \quad \xi \sim \mathcal{N} \bigl(0, \sigma^2_t I \bigl),
    \label{eq:unconditional_langevin}
\end{equation}
where $\sigma_t^2$ is the variance according to a variance schedule and $\gamma$ is the step size\footnote{An linear decay $\tfrac{1}{\sqrt{1-\beta_t}}$ is often also applied to the output $\vx^{t-1}$ for sampling stability.}. This directly corresponds to the noisy energy optimization procedure
\begin{equation}
    \vx^{t-1}=\vx^{t}-\gamma \nabla_{\vx} E_{\theta}(\vx^{t}) + \xi, \quad \xi \sim \mathcal{N} \bigl(0, \sigma^2_t I \bigl).
    \label{eq:ebm_langevin}
\end{equation}
The functional form of \eqn{eq:ebm_langevin} is very similar to \eqn{eqn:opt_approx}, and illustrates how sampling from a diffusion model is similar to optimizing a learned energy function $E_{\theta}(\vx)$ that parameterizes the relative negative log-likelihood of the data density. 

When we train a diffusion model to recover a conditional data density that consists of a single image $\vx_i$, \ie, when we are autoencoding an image given an inferred intermediate latent $\vz$, then the denoising network directly learns an $\epsilon_{\theta}(\vx, \vt, \vz)$ that estimates gradients of an energy function $\nabla_\vx E_\theta(\vx, \vz)$. This energy function has  minimum
\begin{equation}
    \vx_i = \argmin_{\vx} E_\theta(\vx, \vz),
    \label{eqn:image_min}
\end{equation}
as the highest log-likelihood datapoint will be $\vx_i$.
The above equivalence suggests that we may directly use diffusion models to parameterize the unsupervised decomposition of images into the energy functions discussed in \sect{sec:background}.

\begin{figure}[t]
\begin{minipage}{\linewidth}
\begin{algorithm}[H]
    \centering
    \caption{Training Algorithm}
    \label{train_alg}
    \begin{algorithmic}[1]
        \STATE \textbf{Input:} Encoder $\text{Enc}_\cm{\phi}$, denoising model $\epsilon_\theta$, components $K$, data distribution $p_D$
        \WHILE{not converged}
        \STATE $\vx_i \sim p_D$
        \STATE \emph{$\triangleright$ Extract components $\vz_k$ from $\vx_i$}
        \STATE $\vz_1, \ldots,  \vz_K \gets \text{Enc}_\cm{\phi}(\vx_i)$
        \STATE \emph{$\triangleright$ Compute denoising direction} \\
        \STATE $\eps \sim \mathcal{N}(0, 1), t \sim \text{Unif}(\{1, \ldots, T\})$
        \STATE $\vx_i^t = \sqrt{1-\beta_t}\vx_i + \sqrt{\beta_t} \epsilon$ 
        \STATE $\epsilon_{\text{pred}} \gets \sum_k \epsilon_\theta(\vx_i^t, t, \vz_k)$
        
        \STATE \emph{$\triangleright$ Optimize objective $\mathcal{L}_{\text{MSE}}$ wrt $\zeta = \{\phi, \theta\}$:} 
        \STATE $\Delta \zeta \gets \nabla_\zeta  \|\epsilon_{\text{pred}} - \eps\|^2 $
        \ENDWHILE

    \end{algorithmic}
    \label{alg:train}
\end{algorithm}
\end{minipage}
\vspace{-15pt}
\end{figure}

\subsection{Decompositional Diffusion Models}
\label{sect:decomp_diffusion}

In COMET, given an input image $\vx_i$, we are interested in inferring a set of different latent energy functions $E_\theta(\vx, \vz_k)$ such that
\begin{equation*}
    \vx_i = \argmin_\vx \sum_k E_\theta(\vx, \vz_k).
\end{equation*}
Using the equivalence between denoising networks and energy function discussed in~\sect{sect:diffusion_ebm} to recover the desired set of energy functions, we may simply learn a set of different denoising functions to recover an image $\vx_i$ using the objective:
\begin{equation}
 \resizebox{0.9\hsize}{!}{$
 \mathcal{L}_{\text{MSE}}=\left\|\mathbf{\epsilon} - \sum_k \epsilon_\theta\left(\sqrt{1-\beta_t}\vx_i +  \sqrt{\beta_t} \mathbf{\epsilon}, t, \vz_k\right)\right\|_2^2$},
 \label{eq:diffusion_loss}
\end{equation}
where each individual latent $\vz_k$ is inferred by a jointly learned neural network encoder $\text{Enc}_\cm{\phi}(\vx_i)[k]$. We leverage information bottleneck to encourage components to discover independent portions of $\vx_i$ by constraining latent representations $\vz = \{\vz_1, \vz_2, \cdots, \vz_K \}$ to be low-dimensional.
This resulting objective is simpler to train than that of COMET, as it requires only a single step denoising supervision and does not need computation of second-order gradients.

\begin{figure*}[t!]
\begin{center}
\includegraphics[width=\textwidth]{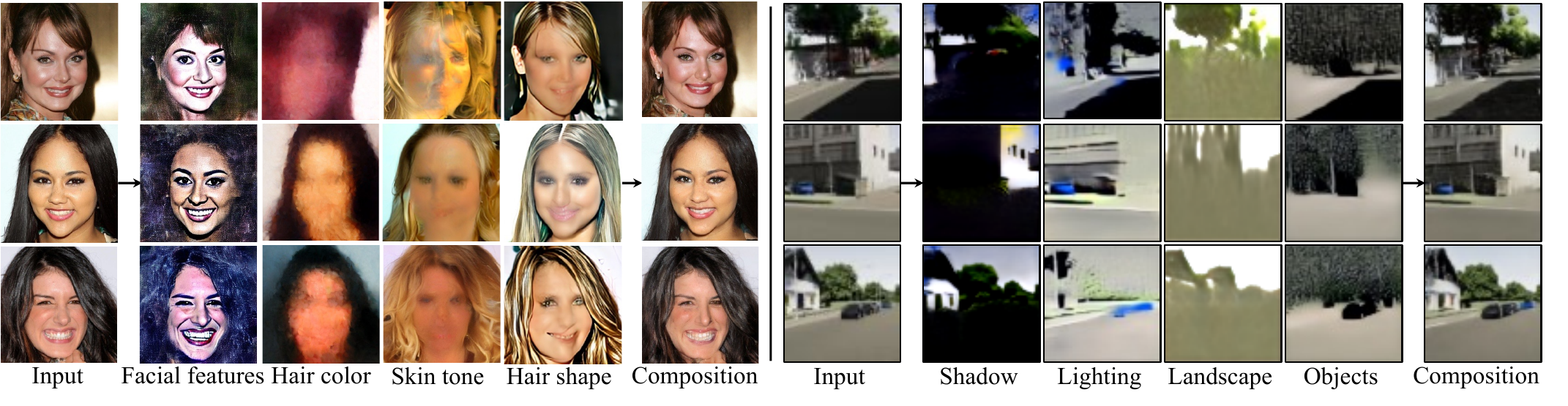}
\end{center}
\vspace{-15pt}
\caption{\textbf{Global Factor Decomposition.} Our method can enable global factor decomposition and reconstruction on CelebA-HQ (\textbf{Left}) and Virtual KITTI $2$ (\textbf{Right}). {\it Note that discovered factors are labeled with posited factors.}}
\label{fig:global_decomp}
\vspace{-10pt}
\end{figure*}

\myparagraph{Reconstruction Training.} As discussed in~\citep{ho2020denoising}, the denoising network $\epsilon_\theta$ may either be trained to directly estimate the starting noise $\epsilon$ or the original image $\vx_i$. These two predictions are functionally identical, as $\eps$ can be directly obtained by taking a linear combination of noisy image $\vx^t$ and $\vx_i$. While standard diffusion training directly predicts $\epsilon$, we find that predicting $\vx_i$ and then regressing $\epsilon$ leads to better performance, as this training objective is more similar to autoencoder training. 

Once we have recovered these denoising functions, we may directly use the noisy optimization objective in \eqn{eq:ebm_langevin} to sample from compositions of different factors. The full training and sampling algorithm for our approach are shown in Algorithm~\ref{train_alg} and Algorithm~\ref{gen_alg} respectively.

\begin{figure}[t]
\begin{minipage}{\linewidth}
\begin{algorithm}[H]
    \centering
    \caption{Image Generation Algorithm}\label{gen_alg}
    
    \begin{algorithmic}[1]
        \STATE \textbf{Input:} Diffusion steps $T$, denoising model $\epsilon_\theta$, latent vectors $\{\vz_1, \hdots, \vz_K\}$, step size $\gamma$
    
        \STATE $\vx^T \sim \mathcal{N}(0, 1)$
        \FOR{$t=T, \ldots, 1$}
            \STATE \emph{$\triangleright$ Sample Gaussian noise}
            \STATE $\xi \sim \mathcal{N}(0, 1)$
            \STATE \emph{$\triangleright$ Compute denoising direction}
            \STATE $\epsilon_{\text{pred}} \gets \sum_k \epsilon_\theta(\vx^{t}, t, \vz_k)$
            \STATE \emph{$\triangleright$ Run noisy gradient  descent}
            \STATE $\vx^{t-1} = \frac{1}{\sqrt{1-\beta_t}} (\vx^t - \gamma \epsilon_{\text{pred}} + \sqrt{\beta_t} \xi) $
        \ENDFOR
\end{algorithmic}
\label{alg:image_gen}
\end{algorithm}
\end{minipage}
\vspace{-5pt}
\end{figure}

\section{Experiments}\label{sec:exp}

In this section, we evaluate the ability of our approach to decompose images. First, we assess decomposition of images into global factors of variation in \sect{sect:global}. We next evaluate decomposition of images into local factors of variation in \sect{sect:local}. We further investigate the ability of decomposed components to recombine across separate trained models in \sect{sect:recomb}. 
Finally, we illustrate how our approach can be adapted to pretrained models in \sect{sect:pretrained}. We use datasets with a degree of consistency among the images, for example aligned face images, to ensure that they have common elements our approach could extract.

\begin{figure}[t!]
\begin{center}
\includegraphics[width=0.48\textwidth]{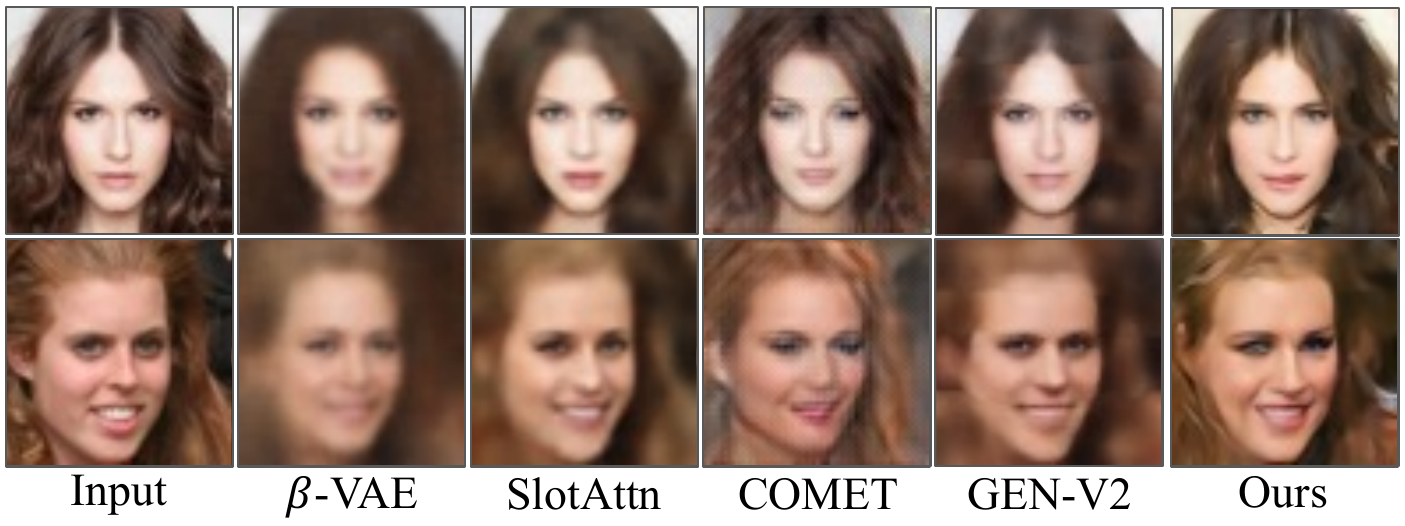}
\end{center}
\vspace{-10pt}
\caption{\textbf{Reconstruction comparison.} Our method can reconstruct input images with a high fidelity on CelebA-HQ.}
\label{fig:celebahq_comparison}
\vspace{-15pt}
\end{figure}

\begin{figure*}[t!]
\begin{center}
\includegraphics[width=\textwidth]{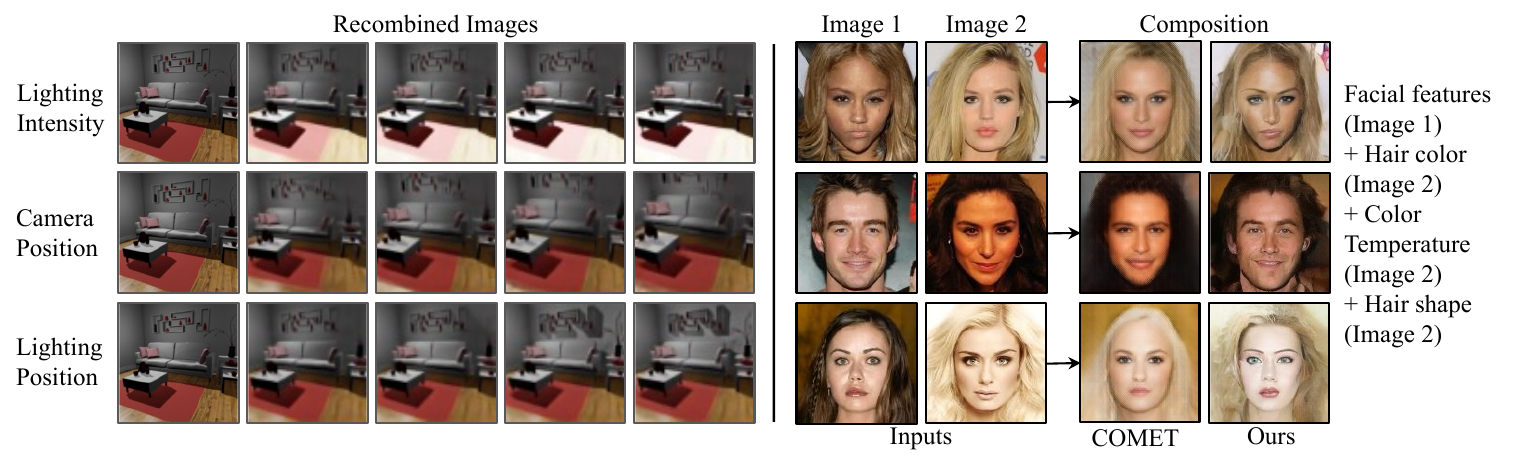}
\end{center}
\vspace{-15pt}
\caption{\textbf{Global Factor Recombination.} Recombination of inferred factors on Falcor3D and CelebA-HQ datasets. In Falcor3D (\textbf{Left}), we show image variations by varying inferred factors such as lighting intensity. In CelebA-HQ (\textbf{Right}), we recombine factors from two different inputs to generate novel face combinations.}
\label{fig:global_recomb}
\vspace{-5pt}
\end{figure*}

\subsection{Quantitative Metrics}
For quantitative evaluation of image quality, we employ Fréchet Inception Distance (FID) ~\cite{heusel2017gans}, Kernel Inception Distance (KID)~\cite{binkowski2018demystifying}, and LPIPS~\cite{zhang2018unreasonable} on images reconstructed from CelebA-HQ~\citep{karras2017progressive}, Falcor3D~\citep{nie2020semi}, Virtual KITTI 2~\citep{cabon2020virtual}, and CLEVR~\citep{Johnson2017CLEVR}. To evaluate disentanglement, we compute MIG~\citep{chen2018isolating} and MCC~\citep{hyvarinen2016unsupervised} on learned latent representation images on the Falcor3D dataset.

\subsection{Global Factors}
\label{sect:global}

Given a set of input images, we illustrate how our unsupervised approach can capture a set of global scene descriptors such as lighting and background and recombine them to construct image variations. We evaluate results in terms of image quality and disentanglement of global components.

\begin{table*}[t]
\small
\setlength{\tabcolsep}{5pt}
\centering
\scalebox{0.9}{\begin{tabular}{lcccccccccccccc}
    \toprule
      {\bf \multirow{2}{*}{Model}} &  \multicolumn{3}{c}{\bf CelebA-HQ} &
      \multicolumn{3}{c}{\bf Falcor3D} &  
        \multicolumn{3}{c}{\bf Virtual KITTI $2$} &
    \multicolumn{3}{c}{\bf CLEVR} \\
      \cmidrule(lr){2-4} \cmidrule(lr){5-7} \cmidrule(lr){8-10} \cmidrule(lr){11-13}
      & FID $\downarrow$ & KID $\downarrow$ & LPIPS $\downarrow$
      & FID $\downarrow$ & KID $\downarrow$ & LPIPS $\downarrow$
      & FID $\downarrow$ & KID $\downarrow$ & LPIPS $\downarrow$ 
      & FID $\downarrow$ & KID $\downarrow$ & LPIPS $\downarrow$\\
      \midrule
       $\beta$-VAE ($\beta = 4$) & $107.29$ & $0.107$ & $0.239$ & $116.96$ & $0.124$ & $0.075$ & $196.68$ & $0.181$ & $0.479$ &
       $\cm{316.64}$ 
       & $\cm{0.383}$
       & $\cm{0.651}$
       \\
       MONet & $35.27$ & $0.030$  & $0.098$ & $69.49$ & $0.067$  & $0.082$ & $67.92$ & $0.043$ & $0.154$ & \cm{$60.74$} & \cm{$0.063$} & \cm{$0.118$} \\
       COMET & $62.64$ & $0.056$  & $0.134$ & $46.37$ & $0.040$ & $0.032$ & $124.57$ & $0.091$ & $0.342$ & \cm{$103.84$} & \cm{$0.119$} & \cm{$0.141$}\\
       Slot Attention & $56.41$ & $0.050$ & $0.154$ & $65.21$ & $0.061$ & $0.079$ & $153.91$ & $0.113$ & $0.207$ & \cm{$27.08$} & \cm{$0.026$} & \cm{$0.031$} \\
       Hessian Penalty & $34.90$ & $0.021$ & -- & $322.45$ & $0.479$ & -- & $116.91$ & $0.084$ & -- & 
       \cm{$25.40$} &
       \cm{$0.016$} &
       \cm{--} \\
       GENESIS-V2 & $41.64$ & $0.035$ & $0.132$ & $ 130.56$ & $0.130$ & $0.097$ & 
       $134.31$ & $0.105$ & $ 0.202$ 
       & \cm{$318.46$} 
       & \cm{$0.403$}
       & \cm{$0.631$}\\

       Ours & $\mathbf{16.48}$ & $\mathbf{0.013}$ & 
       $\mathbf{0.089}$ &
       $\mathbf{14.18}$ & $\mathbf{0.008}$ & 
       $\mathbf{0.028}$ &
       $\mathbf{21.59}$ & $\mathbf{0.008}$ &
       $\mathbf{0.058}$ &
       \cm{$\mathbf{11.49}$} & 
       \cm{$\mathbf{0.011}$} &
       \cm{$\mathbf{0.012}$} \\ %
       
    \bottomrule
\end{tabular}}
\vspace{-5pt}
\caption{\small \textbf{Image Reconstruction Evaluation.} We evaluate the quality of $64 \times 64$ reconstructed images using FID, KID and LPIPS on $10,000$ images from \cm{$4$} different datasets. Our method achieves the best performance.}
\vspace{-5pt}
\label{table:fid}
\end{table*}

\textbf{Decomposition and Reconstruction.} On the left-hand side of  Figure~\ref{fig:global_decomp}, we show how our approach decomposes CelebA-HQ face images into a set of factors. These factors can be qualitatively described as facial features, hair color, skin tone, and hair shape. To better visualize each factor's individual effect, we provide experiments in Figure~\ref{fig:factor_importance_CelebA-HA} where factors are added one at a time to incrementally reconstruct the input image. In addition, we compare our method's performance on image reconstruction against existing baselines in Figure~\ref{fig:celebahq_comparison}. Our method generates better reconstructions than COMET as well as other recent baselines, in that images are sharper and more similar to the input. 

On the right side of Figure~\ref{fig:global_decomp}, we  show how \methodname  infers factors such as shadow, lighting, landscape, and objects on Virtual KITTI $2$. We can further compose these factors to reconstruct the input images, as illustrated in the rightmost column. Comparative decompositions from other methods can be found in~\fig{fig:comet_decomp_celeba}.

We also provide qualitative results to illustrate the effect of number of concepts $K$ on CelebA-HQ and Falcor3D in~\fig{falcor3d_K} and~\fig{celeba_vary_k}, respecticely. As expected, using different $K$ can lead to different sets of decomposed concepts being produced, but certain concepts are learned across different $K$, such as the facial features concepts in~\fig{celeba_vary_k}. 

\begin{table}[t]
\small
\setlength{\tabcolsep}{0.5pt}
\centering
\scalebox{0.9}{\begin{tabular}{lccccc}
 \toprule
 Model & Dim ($D$) & $\beta$ & Decoder Dist. & MIG $\uparrow$ & MCC $\uparrow$ \\
 \midrule
 InfoGAN & $64$ & -- & -- & $2.48\pm1.11$ & $52.67\pm1.91$ \\
  $\beta$-VAE & $64$ & $4$ & Bernoulli & $8.96 \pm 3.53$ & $61.57 \pm 4.09$ \\
 $\beta$-VAE & $64$ & $16$ & Gaussian & $9.33 \pm 3.72$ & $57.28 \pm 2.37$ \\
 $\beta$-VAE & $64$ & $4$ & Gaussian & $10.90\pm3.80$ & $66.08\pm2.00$ \\
 GENESIS-V2* & $128$ & -- & -- & $5.23\pm0.02$ & $63.83\pm0.22$ \\
 MONet & $64$ & -- & -- & $13.94\pm2.09$ & $65.72\pm0.89$ \\
 COMET & $64$ & -- & -- & $19.63\pm2.49$ & $76.55\pm1.35$ \\
  \midrule
 Ours & $32$ & -- & -- & $11.72\pm0.05$ & $57.67\pm0.09$ \\
 Ours & $64$ & -- & -- & $\mathbf{26.45\pm0.16}$ & $\mathbf{80.42\pm0.08}$ \\
  Ours & $128$ & -- & -- & $12.97\pm0.02$ & $80.27\pm 0.17$ \\
  Ours* & $128$ & -- & -- & $16.57\pm0.02$ & $71.19\pm0.15$ \\
 \bottomrule
 \end{tabular}}
 \vspace{-5pt}
\caption{\small \textbf{Disentanglement Evaluation.} Mean and standard deviation of metrics across $3$ random seeds on the Falcor3D dataset. \methodname enables better disentanglement according to $2$ common disentanglement metrics. The asterisk (*) indicates that PCA is applied to project the output dimension to $64$.}
 \label{table:disentangle_quant}
 \vspace{-15pt}
 \end{table}

\textbf{Recombination.} In Figure~\ref{fig:global_recomb}, we explore how factors can be flexibly composed by recombining  decomposed factors from Falcor3D as well as from CelebA-HQ. On the left-hand side, we demonstrate how recombination can be performed on a source image by varying a target factor, such as lighting intensity, while preserving the other factors. This enables us to generate image variations using inferred factors such as lighting intensity, camera position, and lighting position.

On the right-hand side of Figure~\ref{fig:global_recomb}, we show how factors extracted from different faces can be recombined to generate a novel human face that exhibits the given factors. For instance, we can combine the facial features from one person with the hair shape of another to create a new face that exhibits the chosen properties. These results illustrate that our method can effectively disentangle images into global factors that can be recombined for novel generalization.

\begin{figure*}[t!]
\begin{center}
\includegraphics[width=\textwidth]{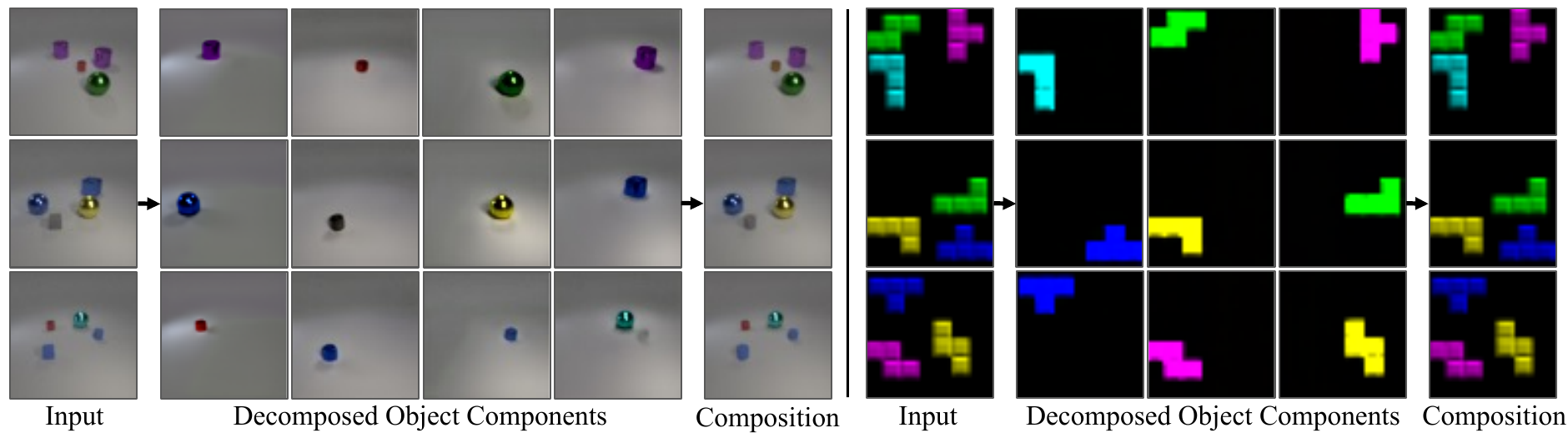}
\end{center}
\vspace{-10pt}
\caption{\textbf{Local Factor Decomposition.} Illustration of object-level decomposition on CLEVR (\textbf{left}) and Tetris (\textbf{right}). Our method can extract individual object components that can be reused for image reconstruction.}
\label{fig:object_decomp}
\vspace{-5pt}
\end{figure*}

\begin{figure*}[t!] %
\begin{center}
\includegraphics[width=\textwidth]{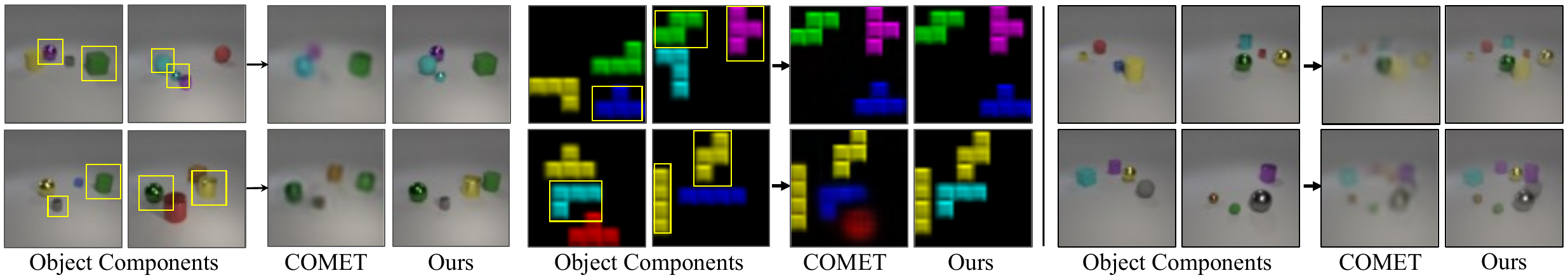}
\end{center}
\vspace{-10pt}
\caption{\textbf{Local Factor Recombination.} We recombine local factors from $2$ images to generate composition of inferred object factors. On both CLEVR and Tetris (\textbf{Left}), we recombine inferred object components in the bounding box to generate novel object compositions. On CLEVR (\textbf{Right}), we compose all inferred factors to generalize up to $8$ objects, though training images only contain $4$ objetcs. }
\label{fig:local_recomb}
\vspace{-5pt}
\end{figure*}

\textbf{Quantitative results.} To quantitatively compare different methods, we evaluate the visual quality of reconstructed images using the decomposed scene factors, as presented in Table~\ref{table:fid}. We observe that our method outperforms existing methods in terms of FID, KID, and LPIPS across datasets, indicating superior image reconstruction quality.

Finally, we evaluate the disentanglement of the given methods on the Falcor3D dataset. As shown in Table~\ref{table:disentangle_quant}, \methodname 
 with dimension  $64$ achieves the best scores across disentanglement metrics, showing its effectiveness in capturing a set of global scene descriptors. In addition, we evaluate our models with different latent dimensions of $32$, $64$, and $128$ to investigate the impact of latent dimension. We find that our method achieves the best performance when using a dimension of $64$. We posit that a smaller dimension may lack the capacity to encode all the information, thus leading to worse disentanglement. A larger dimension may be too large and fail to separate distinct factors. Thus, we apply PCA to project the output dimension $128$ to $64$ (last row), and we observe that it can boost the MIG performance but lower the MCC score. 

\textbf{Diffusion Parameterizations.} We next analyze two choices of diffusion parameterizations for the model, predicting $\vx_0$ or predicting the noise $\epsilon$, in Table~\ref{tab:ablation}. We find that directly predicting the input $\vx_0$ ($3^{\text{rd}}$ \cm{and $6^{\text{th}}$} rows) outperforms the $\epsilon$ parametrization ($1^{\text{st}}$ \cm{and $4^{\text{th}}$} row) \cm{on both CelebA-HQ and CLEVR datasets} in terms of MSE and LPIPS~\citep{zhang2018unreasonable}. This is due to using a reconstruction-based training procedure, as discussed in~\sect{sect:decomp_diffusion}. We also compare using a single component to learn reconstruction ($2^{\text{nd}}$ \cm{and $5^{\text{th}}$} rows) with our method ($3^{\text{rd}}$ \cm{and $6^{\text{th}}$} rows), which uses multiple components for reconstruction. Our method achieves the best reconstruction quality as measured by MSE and LPIPS.

\begin{table}[t]
    \centering
    \setlength{\tabcolsep}{3.5pt}
    \label{table:ablation}
    \resizebox{\linewidth}{!}{\begin{tabular}{lcccccc}
    \toprule  
    \textbf{Dataset} & Multiple  & Predict  & MSE $\downarrow$ & LPIPS $\downarrow$ & FID $\downarrow$ & KID $\downarrow$ \\ 
     & Components & $\vx_0$ &  & & & \\
    \midrule
     & Yes & No & $105.003$ & $0.603$ & \cm{155.46} & \cm{0.141} \\  
    CelebA-HQ & No & Yes & $88.551$ & $0.192$ & \cm{30.10} & \cm{0.022 } \\  
     & Yes & Yes & $\mathbf{76.168}$ & $\mathbf{0.089}$ & 
     \cm{$\mathbf{16.48}$} & \cm{$\mathbf{0.013}$}  \\  
    \midrule %
     & Yes & No & \cm{$56.179$} & \cm{$0.3061$} & \cm{$42.72$} &  \cm{$0.033$} \\  
    CLEVR & No & Yes & \cm{$26.094$} & \cm{$0.2236$} & \cm{$24.27$} & \cm{$0.023$} \\  
     & Yes & Yes & \cm{$\mathbf{6.178}$} & \cm{$\mathbf{0.0122}$} & \cm{$\mathbf{11.54}$} & \cm{$\mathbf{0.010}$}  \\  
    \bottomrule
    \end{tabular}
    }
    \vspace{-5pt}
    \caption{\textbf{Ablations.} We analyze the impact of predicting $\vx_0$ or $\epsilon$, as well as using multiple components or a single component. We compute pixel-wise MSE and LPIPS of reconstructions on \cm{both CLEVR and CelebA-HQ}.}
    \label{tab:ablation}
    \vspace{-20pt}
\end{table}

\begin{figure*}[t!]
\begin{center}
\includegraphics[width=\textwidth]{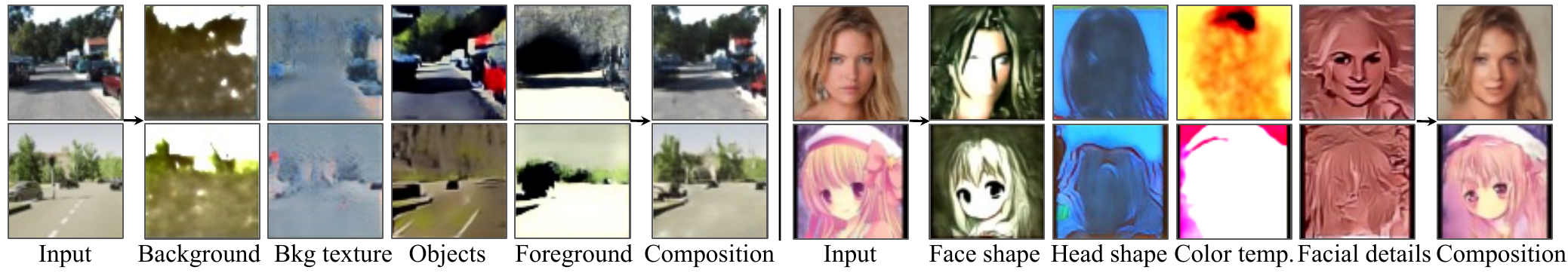}
\end{center}
\vspace{-10pt}
\caption{\textbf{Multi-modal Dataset Decomposition.} We show our method can capture a set of global factors that are shared between hybrid datasets such as KITTI and Virtual KITTI 2 scenes (\textbf{Left}), and CelebA-HQ and Anime faces (\textbf{Right}). {\it Note that discovered factors are labeled with posited factors.}}
\label{fig:multimodel_decomp}
\vspace{-5pt}
\end{figure*}

\begin{figure*}[t!]
\begin{center}
\includegraphics[width=\textwidth ]{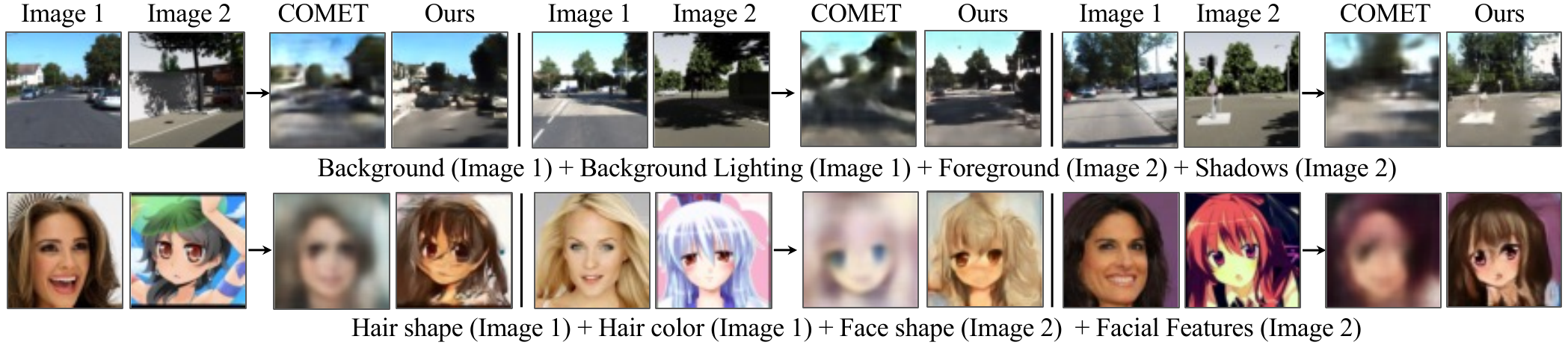}
\end{center}
\vspace{-15pt}
\caption{\textbf{Multi-modal Dataset Recombination.} Our method exhibits the ability to recombine inferred factors from various hybrid datasets. We can recombine different extracted factors to generate unique compositions of KITTI and Virtual KITTI $2$ scenes (\textbf{Top}), and compositions of CelebA-HQ and Anime faces (\textbf{Bottom}). }
\label{fig:multimodel}
\vspace{-15pt}
\end{figure*}

\subsection{Local Factors}
\label{sect:local}

Given an input image with multiple objects, \eg, a purple cylinder and a green cube, we aim to factorize the input image into individual object components using object-level segmentation.

\textbf{Decomposition and Reconstructions.} We qualitatively evaluate local factor decomposition on object datasets such as CLEVR and Tetris in Figure~\ref{fig:object_decomp}. Given an image with multiple objects, our method can both isolate each individual object component as well as faithfully reconstruct the input image using the set of decomposed object factors. Note that since our method does not obtain an explicit segmentation mask per object, it is difficult to quantitatively assess segmentations (though empirically, we found our approach almost always correctly segments objects). We additionally provide results of factor-by-factor compositions, where images are generated by incrementally adding one component at a time, in Figure~\ref{fig:factor_importance_CLEVR}. These mirror the process of adding one object at a time to the scene and demonstrate that our method effectively learns local object-centric representations.

\textbf{Recombination.} To further validate our approach, we show how our method can  recombine local factors from different input images to generate previously unseen image combinations. In Figure~\ref{fig:local_recomb}, we demonstrate how our method utilizes a subset of factors from each image for local factor recombination. On the left-hand side, we present  novel object combinations generated by adding particular factorized energy functions  from two inputs, shown within the bounding boxes, on both the CLEVR and Tetris datasets. On the right-hand side, we demonstrate how our method can recombine all existing local components from two CLEVR images into an unseen combination of $8$ objects, even though each training image only consists of $4$ objects. We illustrate that our approach is highly effective at recombining local factors to create novel image combinations.

\subsection{Cross Dataset Generalization}
\label{sect:recomb}

We next assess the ability of our approach to extract and combine concepts across multiple datasets. We investigate the recombination of factors in multi-modal datasets, as well as the combination of separate factors from distinct models trained on different datasets.

\begin{figure*}[t!] 
\begin{center}
\includegraphics[width=\textwidth]{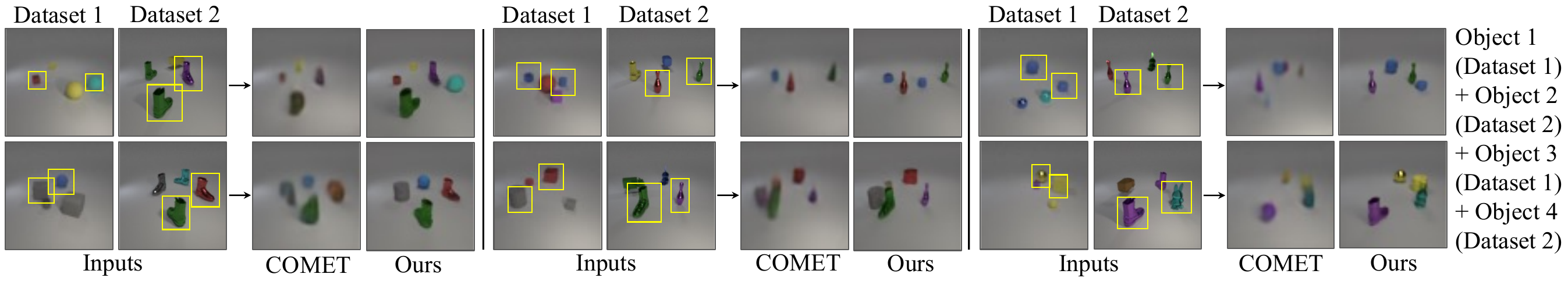}
\end{center}
\vspace{-15pt}
\caption{\textbf{Cross Dataset Recombination.} We further showcase our method's ability to recombine across datasets using $2$ different models that train on CLEVR and CLEVR Toy, respectively. We compose inferred factors as shown in the bounding box from two different modalites to generate unseen compositions.}
\label{fig:clevr_toy}
\vspace{-10pt}
\end{figure*}

\begin{figure}[t!]
\begin{center}

\includegraphics[width=0.5\textwidth]{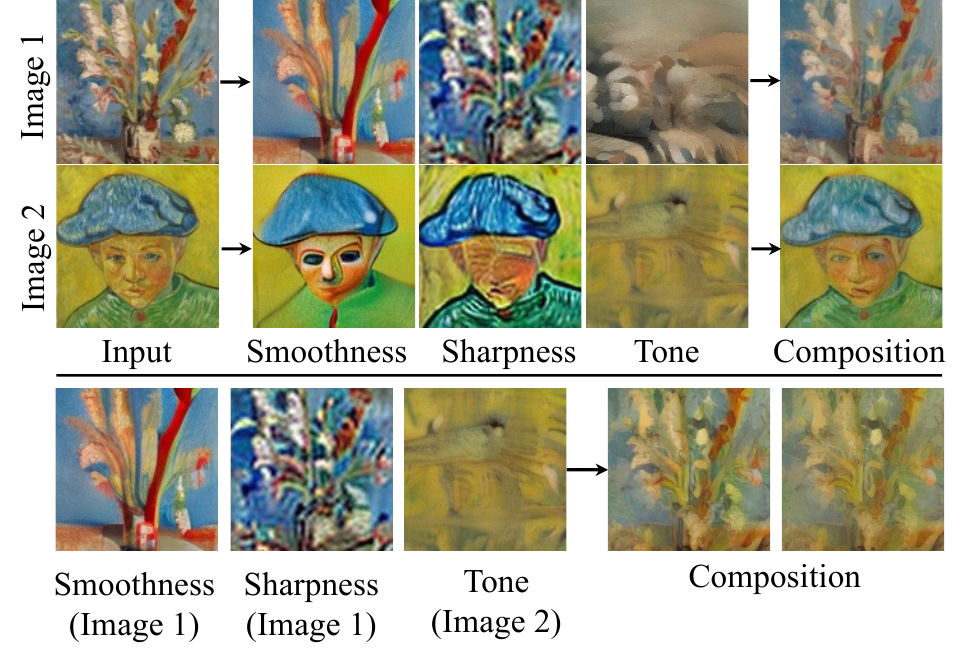}
\end{center}
\vspace{-10pt}
\caption{\textbf{Art Style Decomposition and Recombination.} Illustration of art style decomposition on a Van Gogh painting dataset. Our method can discover art components that capture different facets of the painting content. The discovered factors can be recombined across images to generate novel images.}
\label{fig:art_decomp}
\vspace{-15pt}
\end{figure}

\textbf{Multi-modal Decomposition and Reconstruction.} Multi-modal datasets, such as a dataset containing images from a photorealistic setting and an animated setting, pose a greater challenge for extracting common factors. Despite this, we demonstrate our method's success in this regard in Figure~\ref{fig:multimodel_decomp}. The left-hand side exhibits the decomposition of images from a hybrid dataset comprising KITTI and Virtual KITTI into a set of global factors, such as background, lighting, and shadows. The right-hand side decomposes the two types of faces into a cohesive set of global factors including face shape, hair shape, hair color, and facial details, which can be utilized for reconstruction. This demonstrates our method's effectiveness in factorizing hybrid datasets into a set of factors.

\textbf{Multi-modal Recombination.} Furthermore, we assess the ability of our method to recombine obtained factors across multi-modal datasets, as illustrated in Figure~\ref{fig:multimodel}. 
In the top half, in a hybrid KITTI and Virtual KITTI dataset, we recombine extracted factors from two distinct images to produce novel KITTI-like scenes, for instance incorporating a blue sky background with shadows in the foreground. In the bottom half, we demonstrate our method's ability to  reuse and combine concepts to generate unique anime faces,  combining hair shapes and colors from a human face image with face shape and details from an anime face image.

\textbf{Cross Dataset Recombination.} Given one denoising model $\epsilon_1(x^t, t, z_k)$ trained on the CLEVR dataset and a second denoising model $\epsilon_2(x^t, t, z_n)$ trained on the CLEVR Toy dataset, we investigate  combining local factors extracted from different modalities to generate novel combinations. To compose objects represented by $z_1$ and $z_2$ from one image in CLEVR dataset and objects represented by $z_3$ and $z_4$ from another image in the CLEVR Toy dataset, we sum the predicted individual noise corresponding to ${z_1, z_2, z_3, z_4}$, i.e., $\epsilon_\text{pred} = \epsilon_1(x^t, t, z_1) + \epsilon_1(x^t, t, z_2) + \epsilon_2(x^t, t, z_3) + \epsilon_2(x^t, t, z_4)$, and follow Algorithm 2 to generate a recombined image comprised of objects represented by ${z_1, z_2, z_3, z_4}$. In Figure~\ref{fig:clevr_toy}, our method extracts object components in the bounding box from two images from different datasets, and then further combines them to generate unseen combinations of object components from different models. In Table~\ref{tab:cross}, we provide the FID and KID scores of generated recombinations against the original CLEVR dataset and CLEVR Toy dataset. Our method outperforms COMET on both datasets, indicating the model can obtain better visual quality and more cohesive recombinations.

\subsection{Decomposition with Pretrained Models}
\label{sect:pretrained}

Finally, we illustrate that our approach can adopt pretrained diffusion models as a prior for visual decomposition to avoid training diffusion models from scratch. Specifically, we train the encoder $\text{Enc}_{\phi}$ and finetune Stable Diffusion model $\epsilon_\theta$ together, in the same fashion as shown in Algorithm~\ref{train_alg}. The latent vectors inferred from the encoder are used as conditionings for the Stable Diffusion model to enable image decomposition and composition.

In our experiment, we train our model on a small dataset of $100$ Van Gogh paintings for $1000$ iterations. As shown in Figure~\ref{fig:art_decomp}, our method can decompose such images into a set of distinct factors, such as smoothness, sharpness, and color tone, which can be further recombined to generate unseen compositions like flowers with sharp edges and a yellow tone. Figure~\ref{fig:art_decomp} also shows that our method can use weighted recombination to enhance or reduce individual factors. As an example, we give the tone factor two different weights in the recombination, which results in two images with different extents of yellow tone. This demonstrates that our method can be adapted to existing models efficiently.

\section{Related Work}\label{sec:related}

\myparagraph{Compositional Generation.} Existing work on compositional generation study either modifying the underlying generative process to focus on a set of specifications~\citep{feng2022training,shi2023exploring,cong2023attribute,huang2023composer,garipov2023compositional}, or composing a set of independent models specifying desired constraints~\citep{du2020compositional, liu2021learning,  liu2022compositional,lace,du2023reduce,wang2023concept}. 
Similar to~\citep{du2021unsupervised}, our work aims  discover a set of compositional components from an unlabeled dataset of images which may further be integrated with compositional operations from~\citep{du2023reduce,liu2022compositional}.

\myparagraph{Unsupervised Decomposition.} Unsupervised decomposition focuses on discovering a global latent space which best describes the input space~\citep{Higgins2017Beta,burgess2018understanding,locatello2020disentangling,klindt2021towards,peebles2020hessian,singh2019finegan, preechakul2022diffusion}. In contrast, our approach aims to decompose data into multiple different compositional vector spaces, which allow us to both compose multiple instances of one factor together, as well as compose factors across different datasets. The most similar work in this direction is COMET~\citep{du2021comet}, but unlike COMET we decompose images into a set of different diffusion models, and illustrate how this enables higher fidelity and more scalable image decomposition.

\myparagraph{Unsupervised Object-Centric Learning.} Object-centric learning approaches seek to decompose a scene into objects~\citep{burgess2019monet,greff2019multi, locatello2020objectcentric, engelcke2021genesis, kipf2022conditional,seitzer2022bridging, wang2023slot}, but unlike our method, they are unable to model global factors that collectively affect an image. Furthermore, although some approaches adopt a diffusion model for better local factor decomposition~\citep{jiang2023objectcentric, wu2023slotdiffusion}, they only use the diffusion model as a decoder and still rely on a Slot Attention encoder for decomposition. In contrast, our approach is not limited by a specific encoder architecture because factor discovery is performed by modeling a composition of energy landscapes through the connection between diffusion models and EBMs.

\myparagraph{Diffusion-Based Concept Learning.} Recent diffusion-based approaches often learn to acquire concepts by optimizing token embeddings with a collection of similar images~\citep{lee2023language,chefer2023hidden, avrahami2023break,li2023photomaker, avrahami2023chosen, kumari2023multi, wei2023elite,shah2023ziplora}, and so can be deemed supervised methods. The use of segmentation in decomposition has been explored in other methods, for example using through segmentation masks
~\citep{liu2023ReferringIS, yi2023CVPR, song2023objectstitch} or text captions ~\citep{xu2022GroupViTSS}, while our decomposition approach is completely unsupervised. The most relevant work to ours, ~\citep{liu2023unsupervised} learns to decompose a set of images into a basis set of components using a pretrained text-to-image generative model in an unsupervised manner. However, our work aims to discover components per individual image.

\vspace{-3pt}
\section{Conclusion}

 \looseness=-1
\myparagraph{Limitations.} Our work has several limitations. First, our current approach decomposes images into a fixed number of factors that is specified by the user. While there are cases where the number of components is apparent, in many datasets the number is unclear or may be variable depending on the image. In \sect{sect:additional_experiment}, we study the sensitivity of our approach to the number of components.  We find that we recover duplicate components when the number is too large, and subsets of components when it is too small. A principled approach to determine the ideal number of factors would be an interesting future line of work. In addition, factors discovered by our approach are not guaranteed be distinct from the original image or from each other, and if the latent encoder's embedding dimension is too large, each latent factor may capture the original image itself. Adding explicit regularization to enforce independence between latents would also be a potential area of future research.

\myparagraph{Conclusion.} In this work, we present \methodname and demonstrate its efficacy at decomposing images into both global factors of variation, such as facial expression, lighting, and background, and local factors, such as constituent objects. We further illustrate the ability of different inferred components to compose across multiple datasets and models. We also show that the proposed model can be adapted to existing pretrained models efficiently. We hope that our work inspires future research in unsupervised discovery of compositional representations in images.

\section*{Acknowledgements}
We acknowledge support from NSF grant 2214177; from AFOSR grant FA9550-22-1-0249; from ONR MURI grant N00014-22-1-2740; and from ARO grant W911NF-23-1-0034. Yilun Du is supported by a NSF Graduate Fellowship.

\section*{Impact Statement}

 \looseness=-1
Our proposed approach does not have immediate negative social impact in its current form since  evaluation is carried out on standard datasets. However, our model's ability to generate facial features or objects in a zero-shot manner raises concerns about potential misuse for misinformation. Thus, advocating for responsible usage is crucial. Additionally, like many generative models, there is a risk of introducing biases related to gender or race depending on the training data. Therefore, careful attention must be paid to data collection and curation to mitigate such biases. Our approach can actually benefit many fields such as scene understanding, artwork generation, and robotics.

\bibliography{example_paper,reference}
\bibliographystyle{icml2024}

\newpage
\appendix

\clearpage

\section{Overview}
In this supplementary material, we present additional qualitative results for various domains in \sect{sup:additional_results}. Next, we describe the model architecture for our approach in \sect{sup:model_details}. Finally, we include experiment details on training datasets, baselines, training, and inference in \sect{sup:experiment}.

\section{Additional Results} 
\label{sup:additional_results}
We first provide additional results on global factor decomposition and recombination in \sect{subsec:global_factors}. We then give additional results on object-level decomposition and recombination in \sect{subsec:local_factors}. Finally, we provide more results that demonstrate cross-dataset generalization in \sect{subsec:cross_dataset}.

\subsection{Global Factors}
\label{subsec:global_factors}
\begin{figure*}[t!]
\begin{center}
\includegraphics[width=\textwidth]{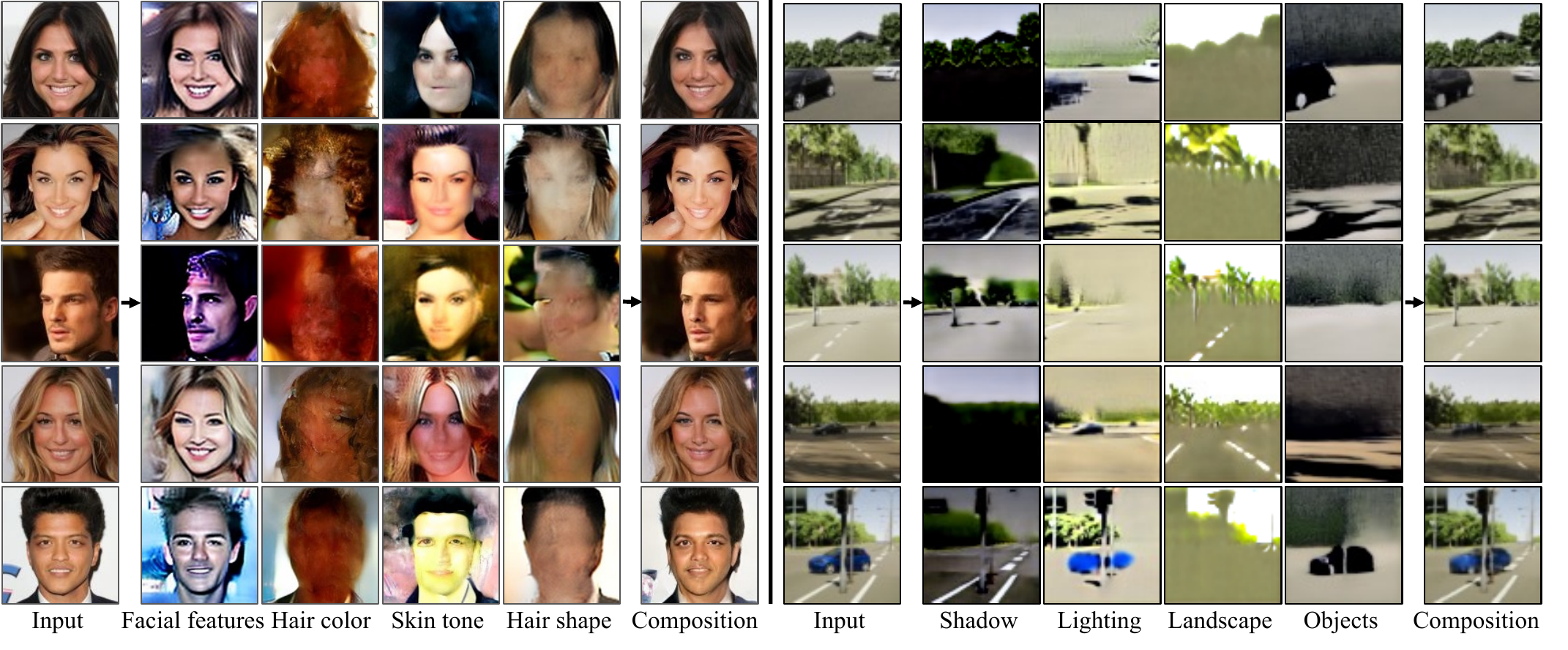} %
\end{center}
\vspace{-15pt}
\caption{\textbf{Global Factor Decomposition.} Global factor decomposition and composition results on CelebA-HQ and Virtual KITTI $2$.  Note that we name inferred concepts for easier understanding.}
\label{fig:global_decomp_extra}
\end{figure*}

\begin{figure*}[t!]
\begin{center}
\includegraphics[width=\textwidth]{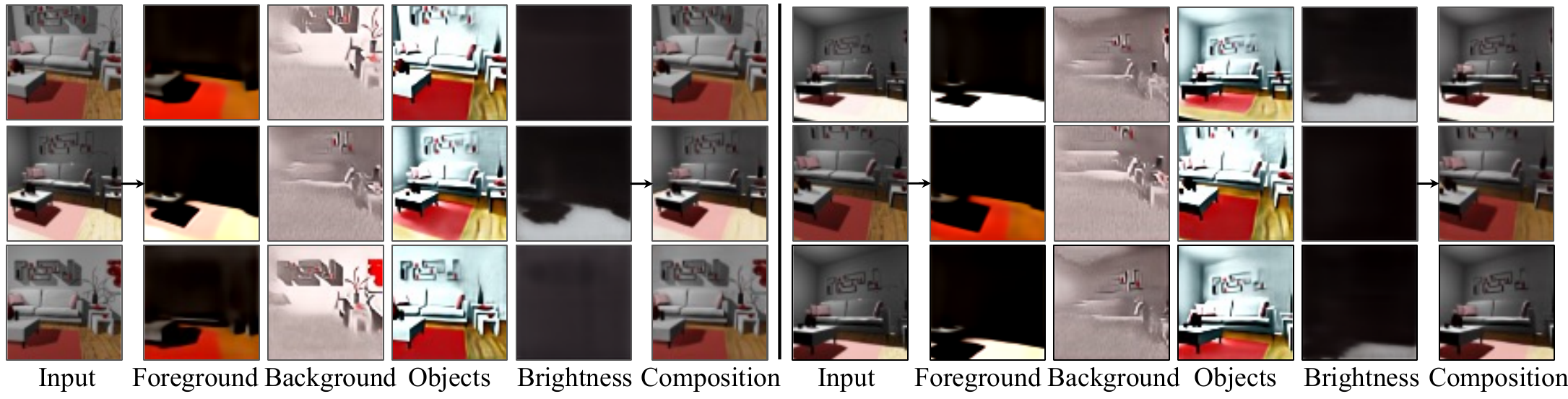}
\end{center}
\vspace{-15pt}
\caption{\textbf{Global Factor Decomposition.} Global factor decomposition and composition results on Falcor3D. Note that we name inferred concepts for easier understanding.}
\label{fig:global_decomp_sup}
\end{figure*}

\textbf{Decomposition and Reconstruction.} In Figure~\ref{fig:global_decomp_extra}, we present supplemental image generations that demonstrate our approach's ability to capture global factors across different domains, such as human faces and scene environments. The left side of the figure displays how our method can decompose images into global factors like facial features, hair color, skin tone, and hair shape, which can be further composed to reconstruct the input images. On the right, we show additional decomposition and composition results using Virtual KITTI $2$ images. Our method can effectively generate clear, meaningful global components from input images. In Figure~\ref{fig:global_decomp_sup}, we show decomposition and composition results on Falcor3D data. Through unsupervised learning, our approach can accurately discover a set of global factors that include foreground, background, objects, and lighting.

\begin{figure*}[t!]
\begin{center}
\includegraphics[width=\textwidth]{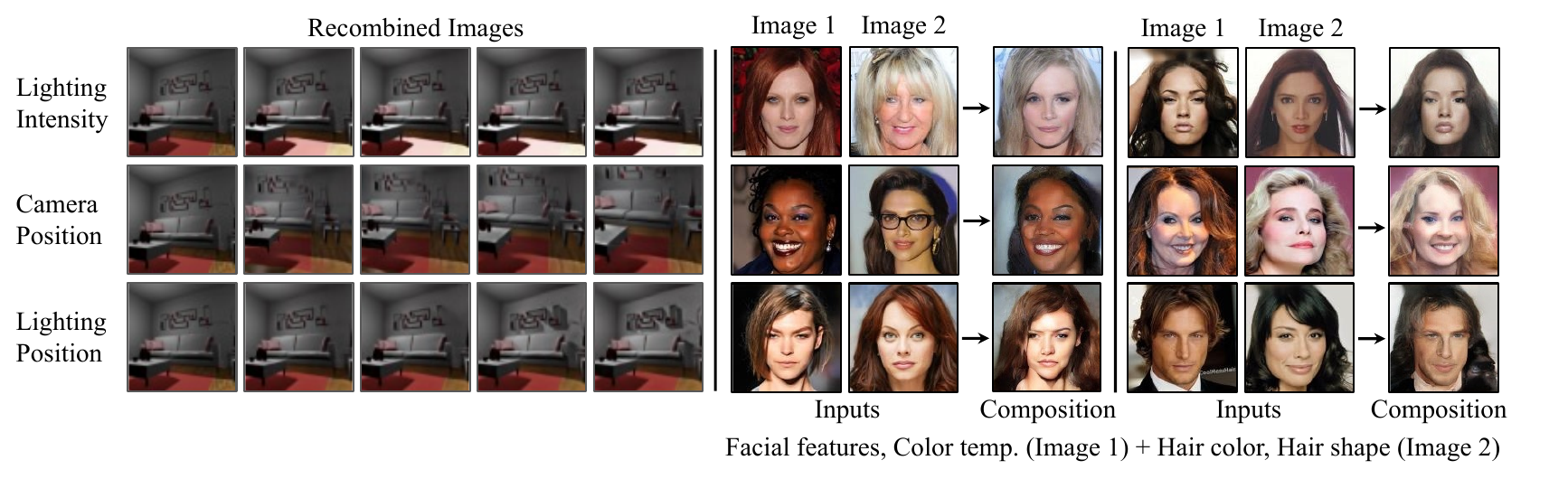}
\end{center}
\vspace{-15pt}
\caption{\textbf{Global Factor Recombination.} Recombination of inferred factors on Falcor3D and CelebA-HQ datasets. In Falcor3D (\textbf{Left}), we show image variations by varying inferred factors such as lighting intensity. In CelebA-HQ (\textbf{Right}), we recombine factors from two different inputs to generate novel face combinations.}
\label{fig:global_recomb_sup}
\end{figure*}

\textbf{Recombination.} Figure~\ref{fig:global_recomb_sup} showcases our approach's ability to generate novel image variations through recombination of inferred concepts. The left-hand side displays results of the recombination process on Falcor3D data, with variations on lighting intensity, camera position, and lighting position. On the right-hand side, we demonstrate how facial features and skin tone from one image can be combined with hair color and hair shape from another image to generate novel human face image combinations. Our method demonstrates great potential for generating diverse and meaningful image variations through concept recombination.

\subsection{Local Factors}
\label{subsec:local_factors}

\begin{figure*}[t!]
\begin{center}
\includegraphics[width=\textwidth]{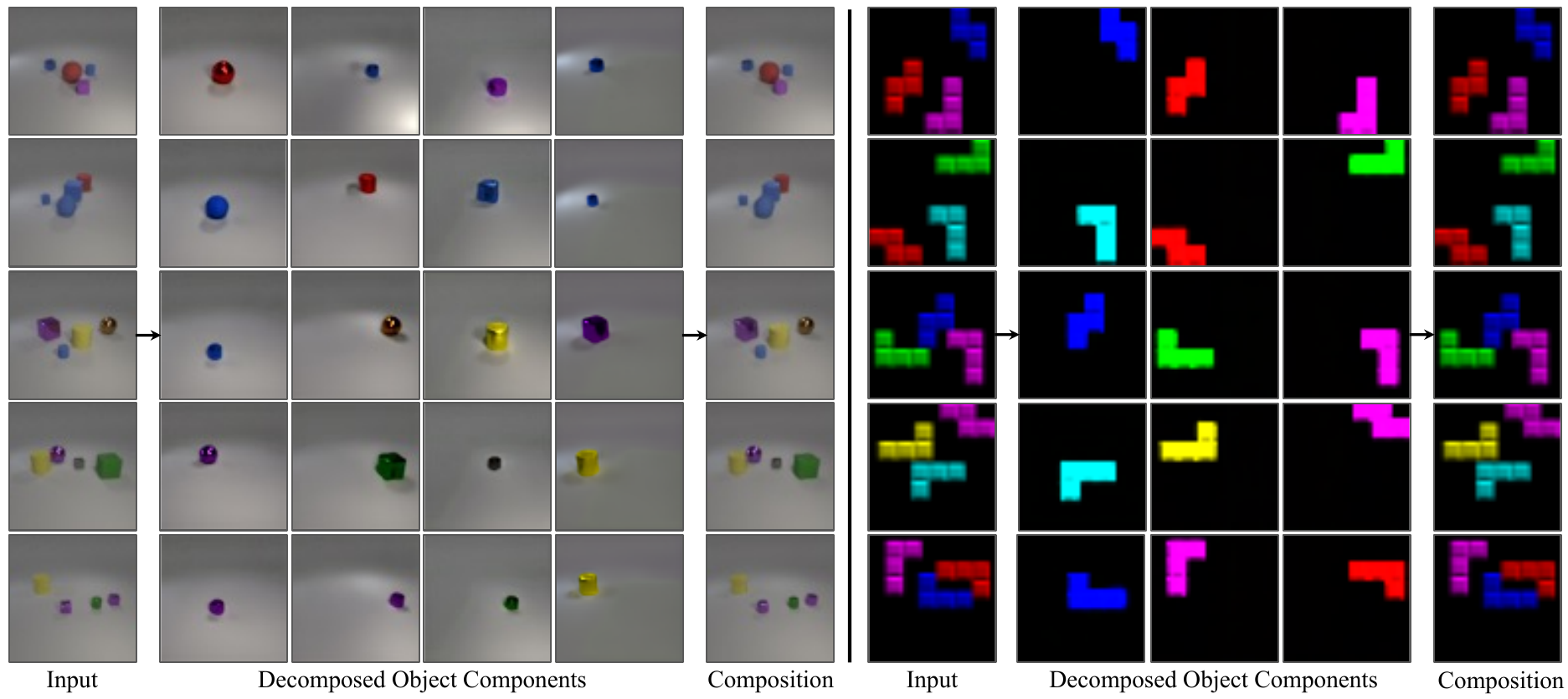}
\end{center}
\vspace{-15pt}
\caption{\textbf{Local Factor Decomposition.} Object-level decompositions results on CLEVR (\textbf{left}) and Tetris (\textbf{right}).}
\label{fig:object_decomp_extra}
\end{figure*}

\textbf{Decomposition and Reconstruction.} We present additional results for local scene decomposition in Figure~\ref{fig:object_decomp_extra}. Our proposed method successfully factorizes images into individual object components, as demonstrated in both CLEVR (\textbf{Left}) and Tetris (\textbf{Right}) object images. Our approach also enables the composition of all discovered object components for image reconstruction.

\begin{figure*}[t!] %
\begin{center}
\includegraphics[width=\textwidth]{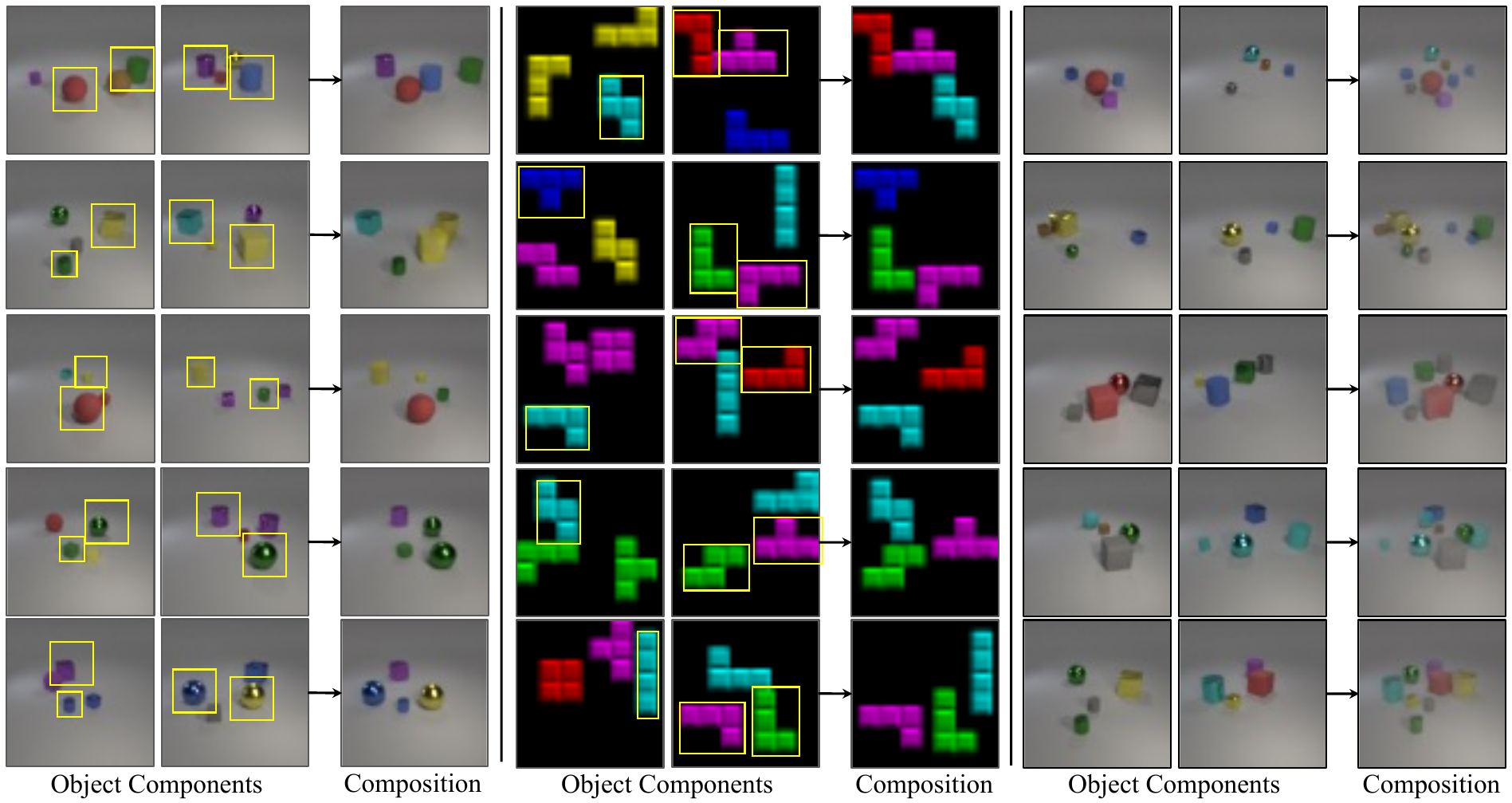}
\end{center}
\vspace{-15pt}
\caption{\textbf{Local Factor Recombination.} Recombination results using object-level factors from different images.}
\label{fig:local_recomb_extra}
\end{figure*}

\textbf{Recombination.} We demonstrate the effectiveness of our approach for recombination of local scene descriptors extracted from multi-object images such as CLEVR and Tetris. As shown in Figure \ref{fig:local_recomb_extra}, our method is capable of generating novel combinations of object components by recombining the extracted components (shown within bounding boxes for easy visualization). Our approach can effectively generalize across images to produce unseen combinations.

\subsection{Cross Dataset Generalization}
\label{subsec:cross_dataset}

We investigate the recombination of factors inferred from multi-modal datasets, and the combination of separate factors extracted from distinct models trained on different datasets.

\begin{figure*}[t!]
\begin{center}
\includegraphics[width=\textwidth ]{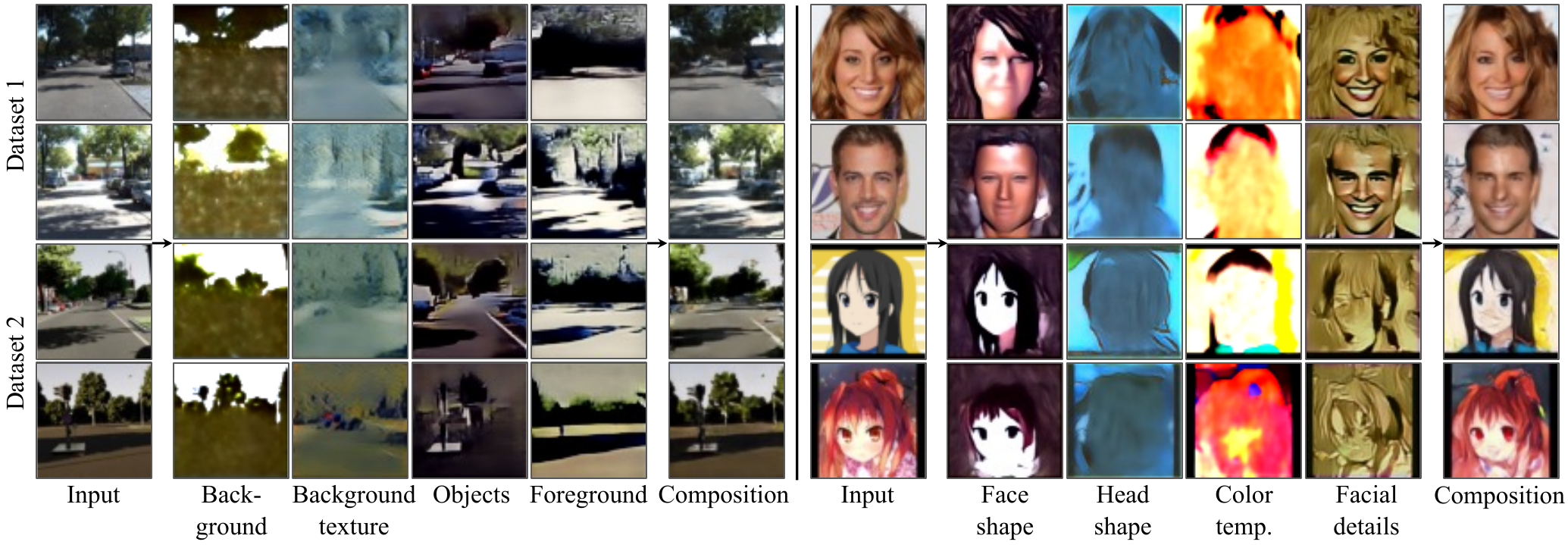}
\end{center}
\vspace{-10pt}
\caption{\textbf{Multi-modal Dataset Decomposition.} Multi-model decomposition and composition results on hybrid datasets such as KITTI and Virtual KITTI 2 scenes (\textbf{Left}), and CelebA-HQ and Anime faces (\textbf{Right}). The top 2 images are of the first dataset, and the bottom 2 images are of the second dataset. Inferred concepts are named for better understanding.}
\label{fig:multimodal_decomp_sup}
\end{figure*}

\textbf{Multi-modal Decomposition and Reconstruction.} We further demonstrate our method's capability to infer a set of factors from multi-modal datasets, \ie, a dataset that consists of different types of images. On the left side of Figure \ref{fig:multimodal_decomp_sup}, we provide additional results on a multi-modal dataset that consists of KITTI and Virtual KITTI $2$. On the right side, we show more results on a multi-modal dataset that combines both CelebA-HQ and Anime datasets.

\begin{figure*}[t!]
\begin{center}
\includegraphics[width=\textwidth ]{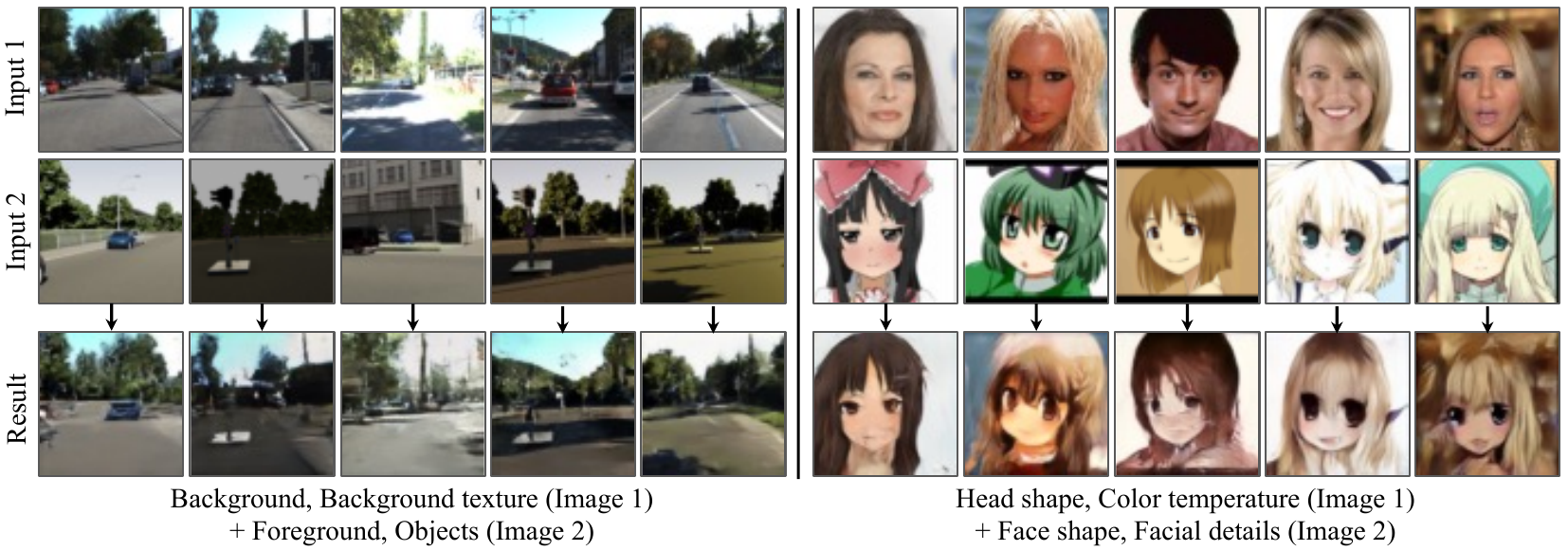}
\end{center}
\vspace{-10pt}
\caption{\textbf{Multi-modal Dataset Recombination.} Recombinations of inferred factors from hybrid datasets. We recombine different extracted factors to generate unique compositions of KITTI and Virtual KITTI $2$ scenes (\textbf{Left}), and compositions of CelebA-HQ and Anime faces (\textbf{Right}). }
\label{fig:multimodal_extra}
\end{figure*}

\textbf{Multi-modal Recombination.} In Figure~\ref{fig:multimodal_extra}, we provide additional recombination results on the two multi-modal datasets of KITTI and Virtual KITTI $2$ on the left hand side of the Figure, and CelebA-HQ and Anime datasets on the right hand side of the Figure.

\begin{figure*}[t!] 
\begin{center}
\includegraphics[width=\textwidth]{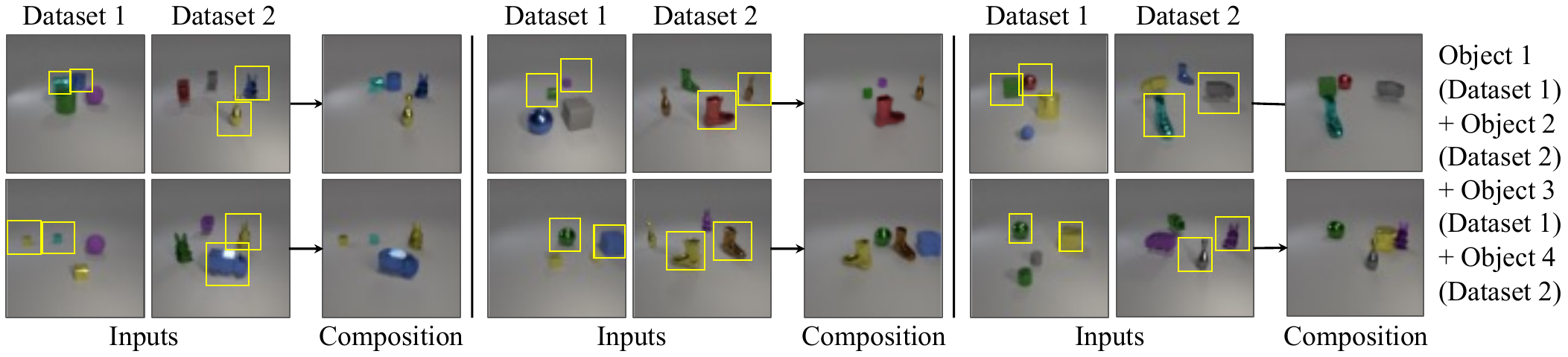}
\end{center}
\vspace{-10pt}
\caption{\textbf{Cross Dataset Recombination.} We further showcase our method's ability to recombine across datasets using $2$ different models that train on CLEVR and CLEVR Toy, respectively. We compose inferred factors as shown in the bounding box from two different modalites to generate unseen compositions.}
\label{fig:clevr_toy_sup}
\end{figure*}

\textbf{Cross Dataset Recombination.} We also show more results for factor recombination across two different models trained on different datasets. In Figure~\ref{fig:clevr_toy_sup}, we combine inferred object components from a model trained CLEVR images and components from a model trained on CLEVR Toy images. Our method enables novel recombinations of inferred components from two different models.

\section{Additional Experiments} %
\label{sect:additional_experiment}

\begin{figure}[t!] 
\begin{center}
\includegraphics[width=0.48\textwidth]{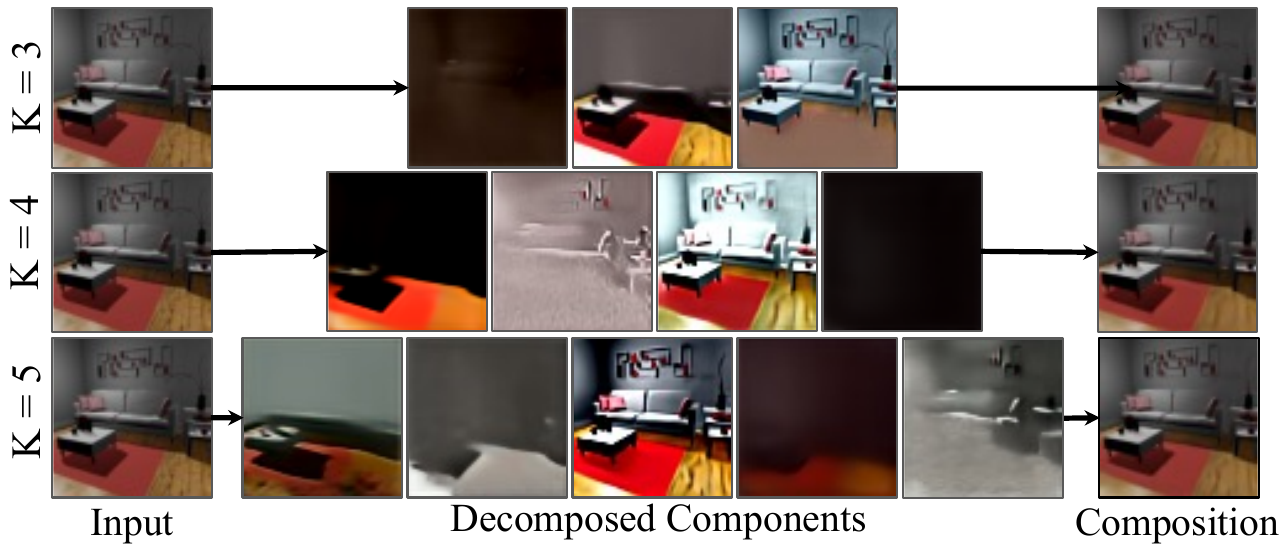}
\end{center}
\vspace{-15pt}
\caption{Decomp Diffusion trained on Falcor3D dataset with varying number of components $K=3, 4,$ and $5$}
\label{falcor3d_K}
\end{figure}

\begin{figure}[t!] 
\begin{center}
\includegraphics[width=0.48\textwidth]{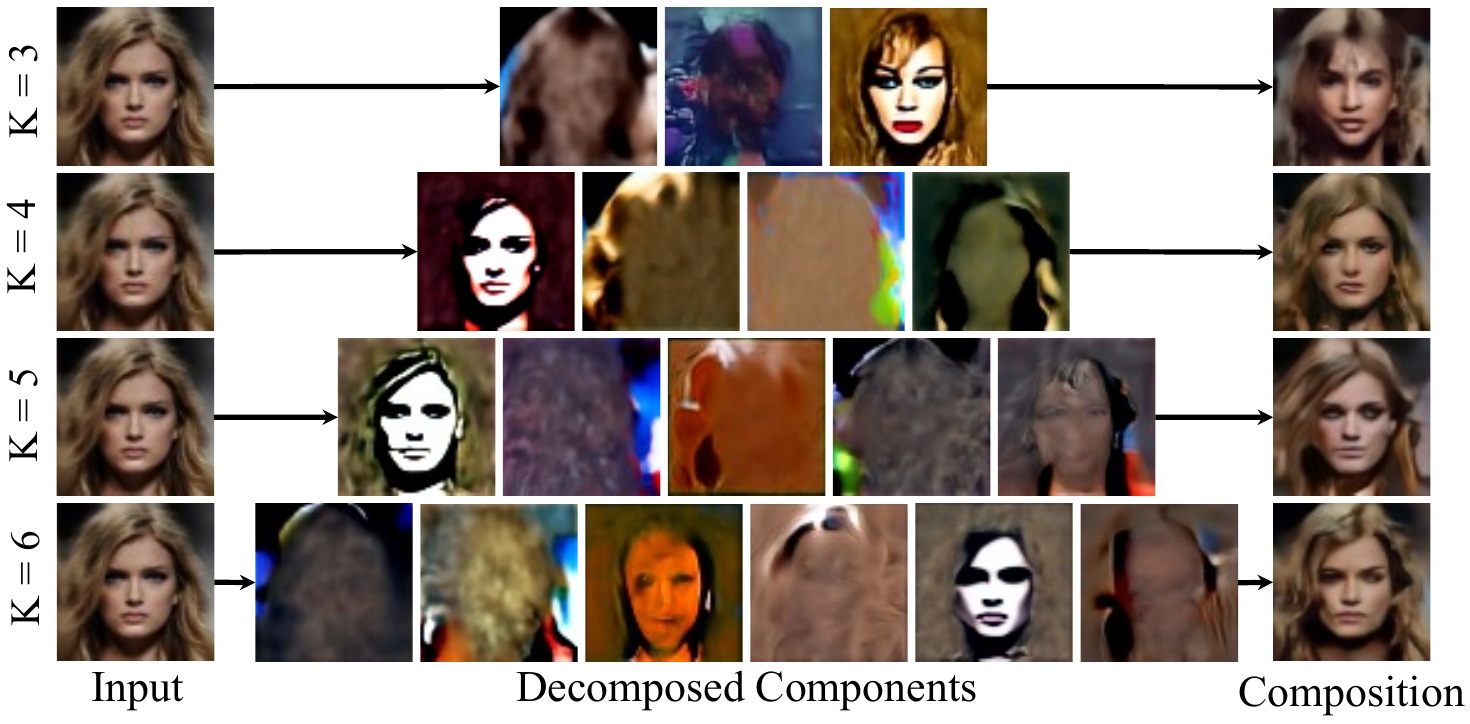}
\end{center}
\vspace{-15pt}
\caption{Decomp Diffusion trained on CelebA-HQ with varying number of components $K=3, 4, 5$, and $6$}
\label{celeba_vary_k}
\end{figure}

\noindent\textbf{Impact of the Number of Components $K$}. We provide qualitative comparisons on the number of components $K$ used to train our models in Figure \ref{falcor3d_K} and Figure \ref{celeba_vary_k}.

\begin{figure*}[t!] 
\begin{center}
\includegraphics[width=\textwidth]{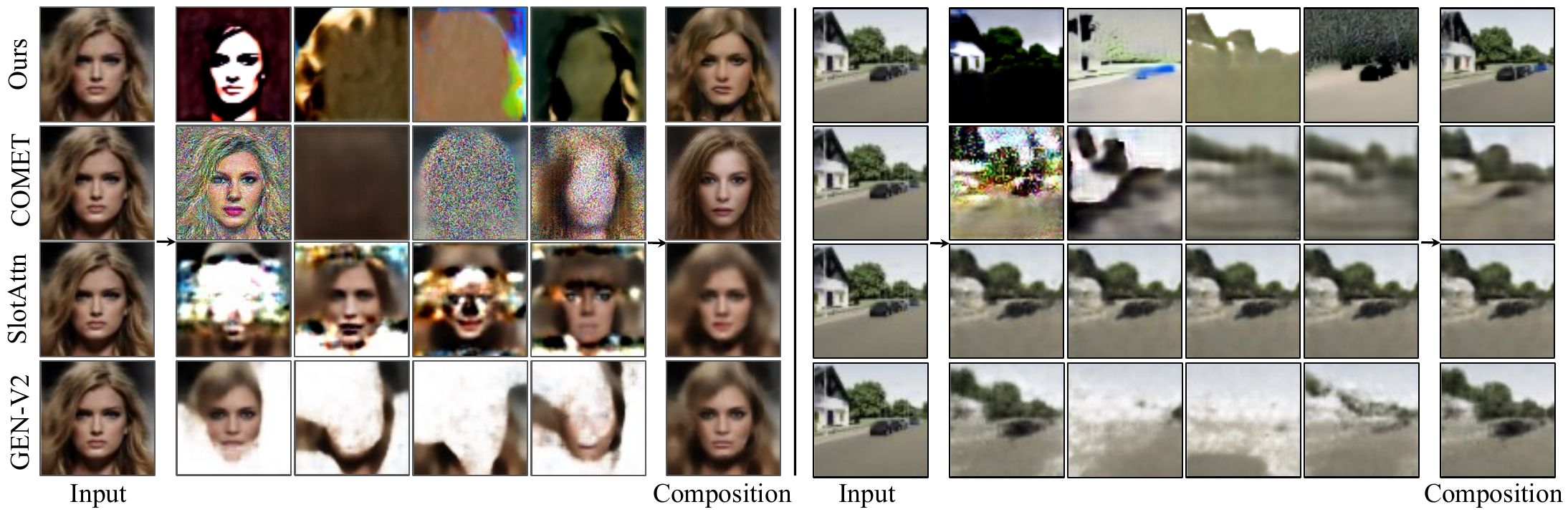}

\end{center}
\vspace{-10pt}
\caption{\textbf{Qualitative comparisons on CelebA-HQ and VKITTI datasets}. Decomposition results on CelebA-HQ (\textbf{Left}) and Virtual KITTI 2 (\textbf{Right}) on benchmark object representation methods. Compared to our method, COMET generates noisy components and less accurate reconstructions. SlotAttention may produce identical components, and it and GENESIS-V2 cannot disentangle global-level concepts. }
\label{fig:comet_decomp_celeba}
\end{figure*}

\begin{figure}[t!] 
\begin{center}
\includegraphics[width=0.47\textwidth]{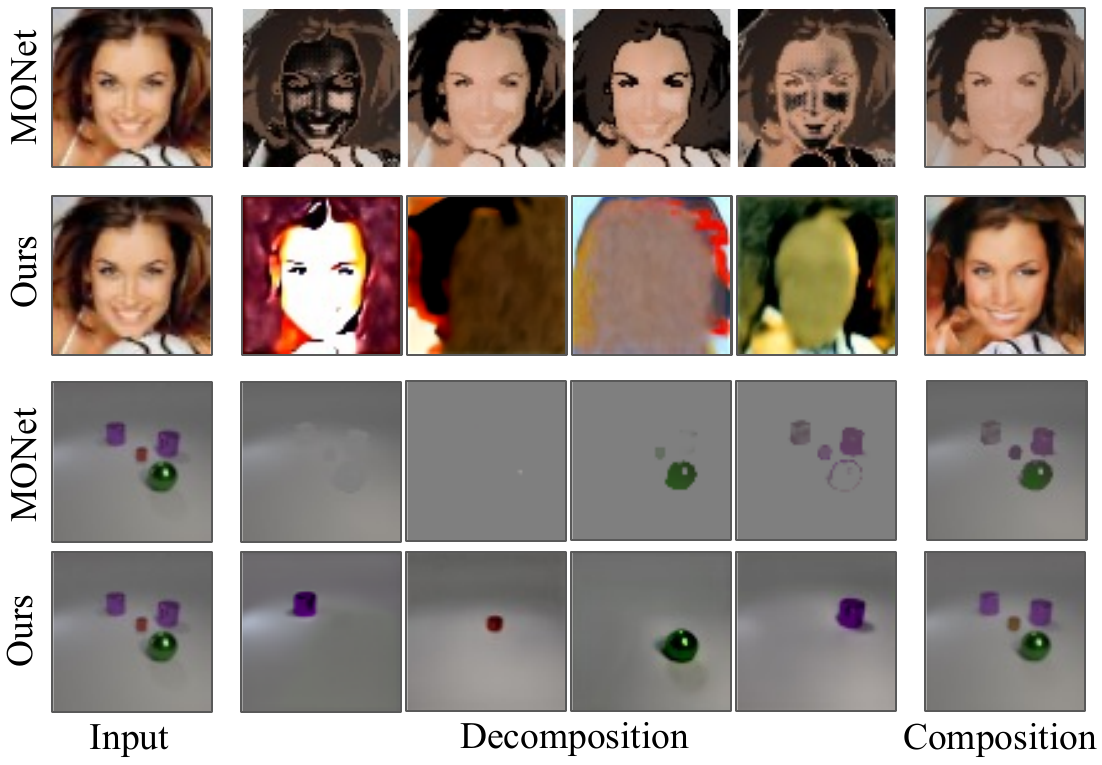}

\end{center}
\vspace{-10pt}
\caption{\cm{\textbf{Decomposition comparisons on CelebA-HQ and CLEVR datasets}. We provide qualitative comparisons on decomposition between MONet and our method.  Our method can decompose images into factors that are more visually diverse and meaningful, while MONet may fail to disentangle factors.} }
\label{fig:monet_decomp}
\end{figure}

\noindent\textbf{Decomposition Comparisons}. We provide qualitative comparisons of decomposed concepts
in Figure \ref{fig:comet_decomp_celeba} \cm{and Figure~\ref{fig:monet_decomp}}.

\begin{figure*}[t!] 
\begin{center}
\includegraphics[width=\textwidth]{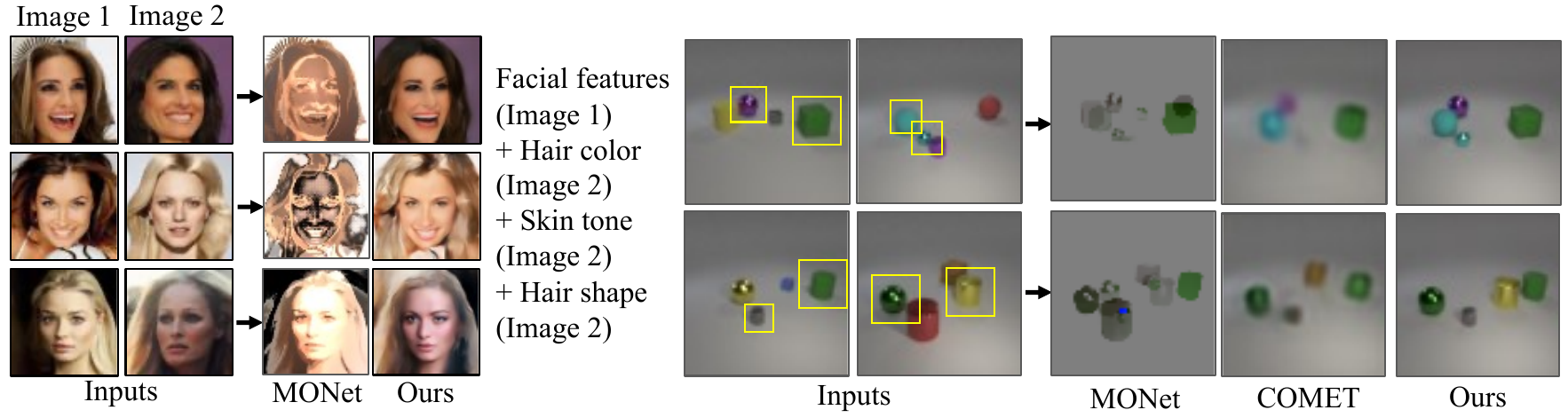}

\end{center}
\vspace{-10pt}
\caption{\cm{\textbf{Recombination comparisons on CelebA-HQ and CLEVR with MONet}. We further compare with MONet on recombination. Our method outperforms MONet by generating correct recombinations results.}}
\label{fig:monet_recombo}
\end{figure*}

\begin{figure*}[t!] 
\begin{center}
\includegraphics[width=\textwidth]{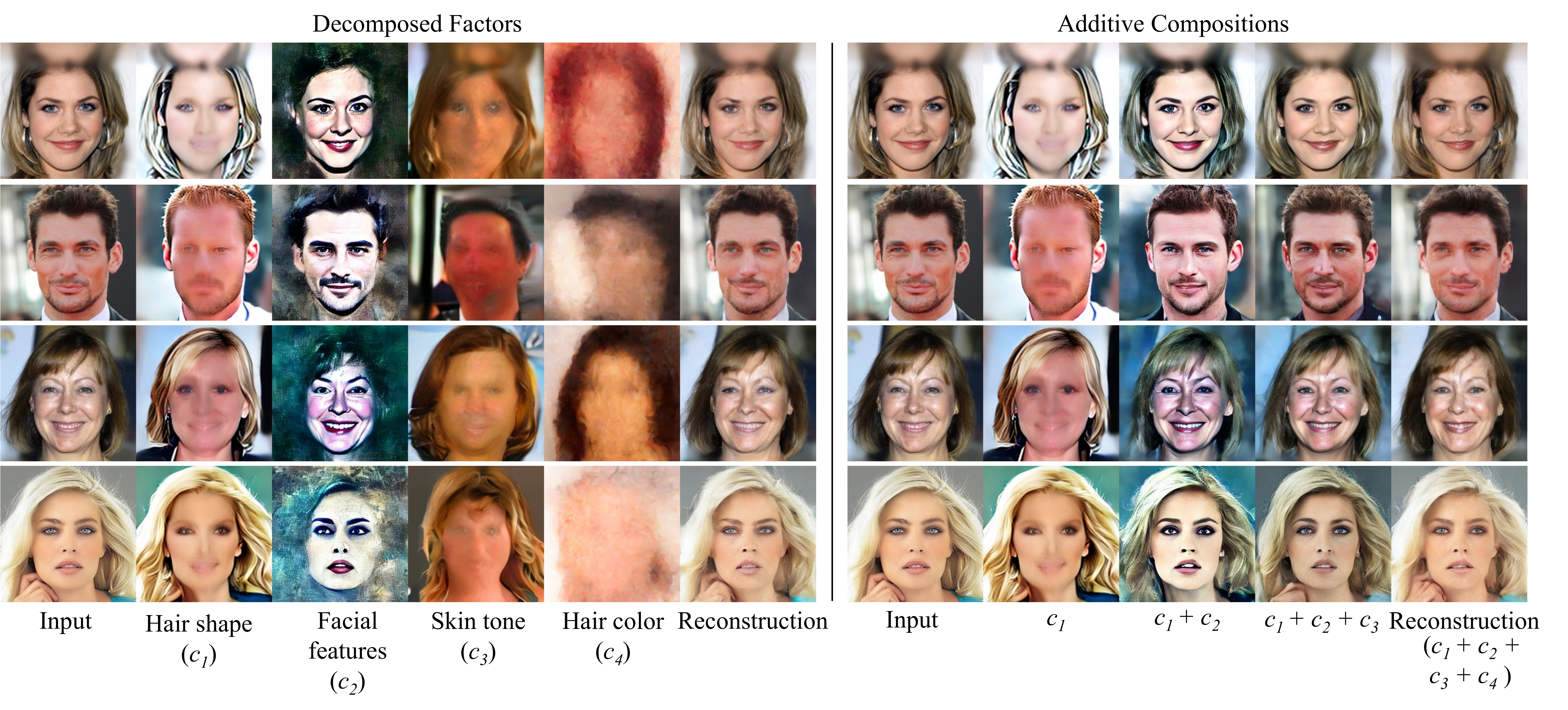}
\end{center}
\vspace{-15pt}
\caption{\textbf{Additive Factors Composition on CelebA-HQ.} On the left, we show decomposed components on CelebA-HQ images with inferred labels. On the right, we present compositions generated by adding one factor at a time to  observe the information learned by each component.}
\label{fig:factor_importance_CelebA-HA}
\end{figure*}

\begin{figure}[t!] 
\begin{center}
\includegraphics[width=0.45\textwidth]{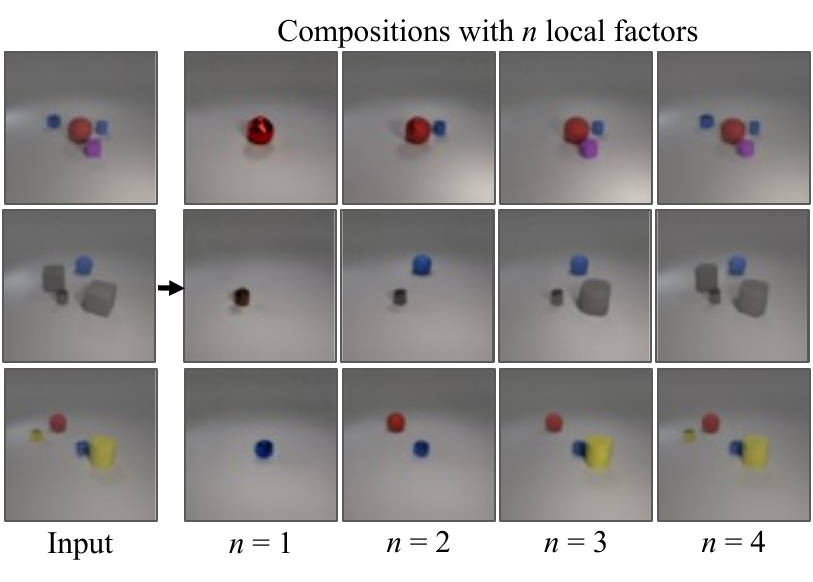}
\end{center}
\vspace{-15pt}
\caption{\textbf{Additive Factors Composition on CLEVR.} We demonstrate that each decomposed object factor can be additively composed to reconstruct the original input image.}
\label{fig:factor_importance_CLEVR}
\end{figure}

\noindent\textbf{Factor Semantics}. To visualize the impact of each decomposed factor, in Figure~\ref{fig:factor_importance_CelebA-HA}, we present composition results produced by incrementally adding components. On the left-hand side, we show the factors discovered for each input image. On the right-hand side, we iteratively add one factor to our latent vector subset and generate the composition results. We see that composition images steadily approach the original input image with the addition of each component. We provide similar additive composition results on the CLEVR dataset in Figure~\ref{fig:factor_importance_CLEVR}. Our method can iteratively incorporate each object represented by the learned local factors until it reconstructs the original image's object setup.

\begin{figure}[t!] 
\begin{center}
\includegraphics[width=0.5\textwidth]{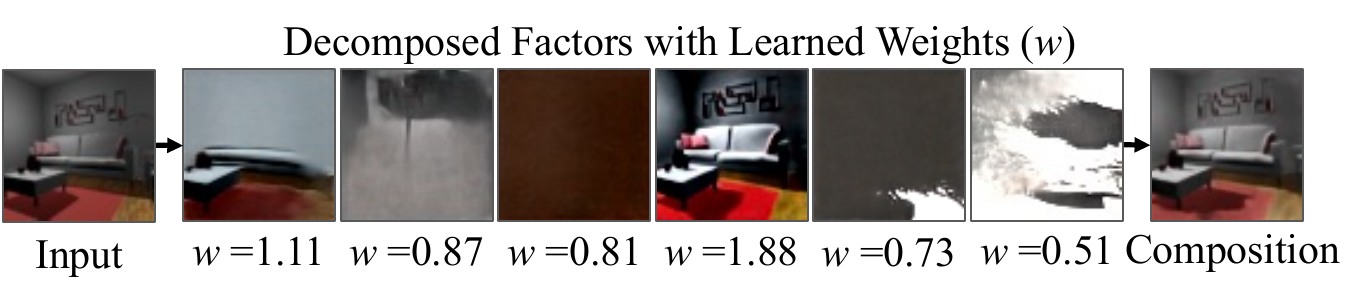}
\end{center}
\vspace{-15pt}
\caption{\textbf{Systematic Selection of Latent Set Size.} We can optionally learn weights for latent components during training. This approach is helpful for automatically choosing the number of components, as we can remove the most insignificant latent components based on their weights.}
\label{fig:latent_set_size_choice}
\end{figure}

\noindent\textbf{Systematic Selection of Latent Set Size}. As a proxy for determining the optimal number of components for decomposition, we conduct reconstruction training by employing a weighted combination of $K$ components, where $K$ is sufficiently large and the weights are learned, rather than simply averaging $K$ components. Subsequently, we utilize the weight values to identify some $K'$ components that were less significant, indicated by their lower weights. The remaining $K - K'$ components may offer a more suitable fit for the dataset. In Figure~\ref{fig:latent_set_size_choice}, we used $K=6$ and found that model learns to differentiate the importance of each component.

\begin{figure}[t!] 
\begin{center}
\includegraphics[width=0.5\textwidth]{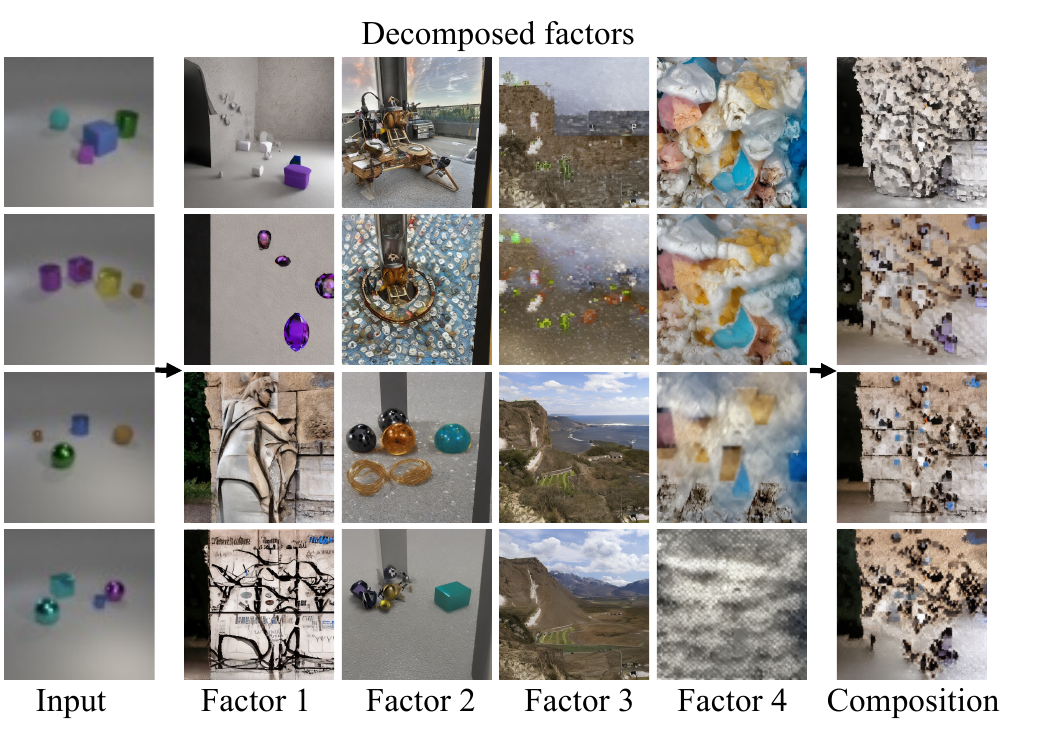}
\end{center}
\vspace{-15pt}
\caption{\textbf{One-Shot Decomposition using~\cite{liu2023unsupervised}.} The method fails to decompose objects in the input training image.}
\label{fig:liu_decomp}
\end{figure}

\noindent\textbf{One-Shot Decomposition with Liu et al.}.

We experiment with using the method from Liu et al. 2023 [4] on a single training image to decompose CLEVR. As shown in Figure \ref{fig:liu_decomp}, since the method only optimizes the word embedding in the text encoder without updating the U-net, it does not generate objects that look similar to the training set. This suggests that the pretrained Stable Diffusion model does not always give faithful priors for factor representation learning tasks.

\begin{figure*}[t!] 
\begin{center}
\includegraphics[width=1\textwidth]{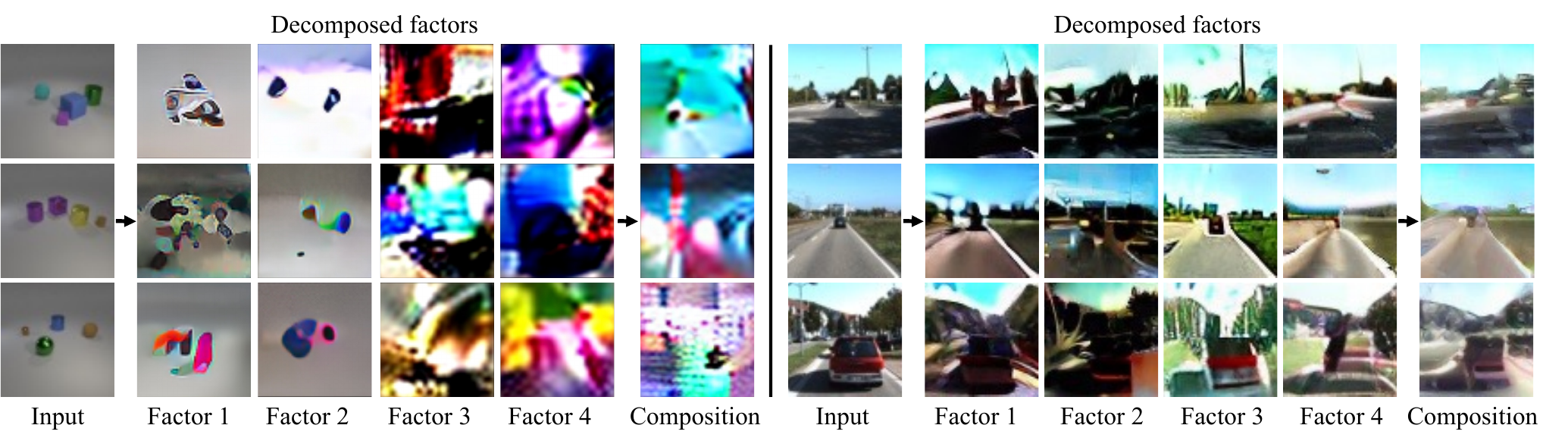}
\end{center}
\vspace{-15pt}
\caption{\textbf{Decomposition with Pretrained Stable Diffusion.} We find that applying our approach with pre-trained Stable Diffusion model doesn't not help find meaningful factors on both CLEVR and KITTI datasets.}
\label{fig:pretrained_sd_decomps}
\end{figure*}

\noindent\textbf{Decomposition with Pretrained Stable Diffusion}
We test a variant of our approach with pretrained Stable Diffusion without fine-tuning on the KITTI and CLEVR datasets, shown in \ref{fig:pretrained_sd_decomps}. We can see that just using the pretrained model did not help find meaningful factors. 

\begin{figure}[t!] 
\begin{center}
\includegraphics[width=0.5\textwidth]{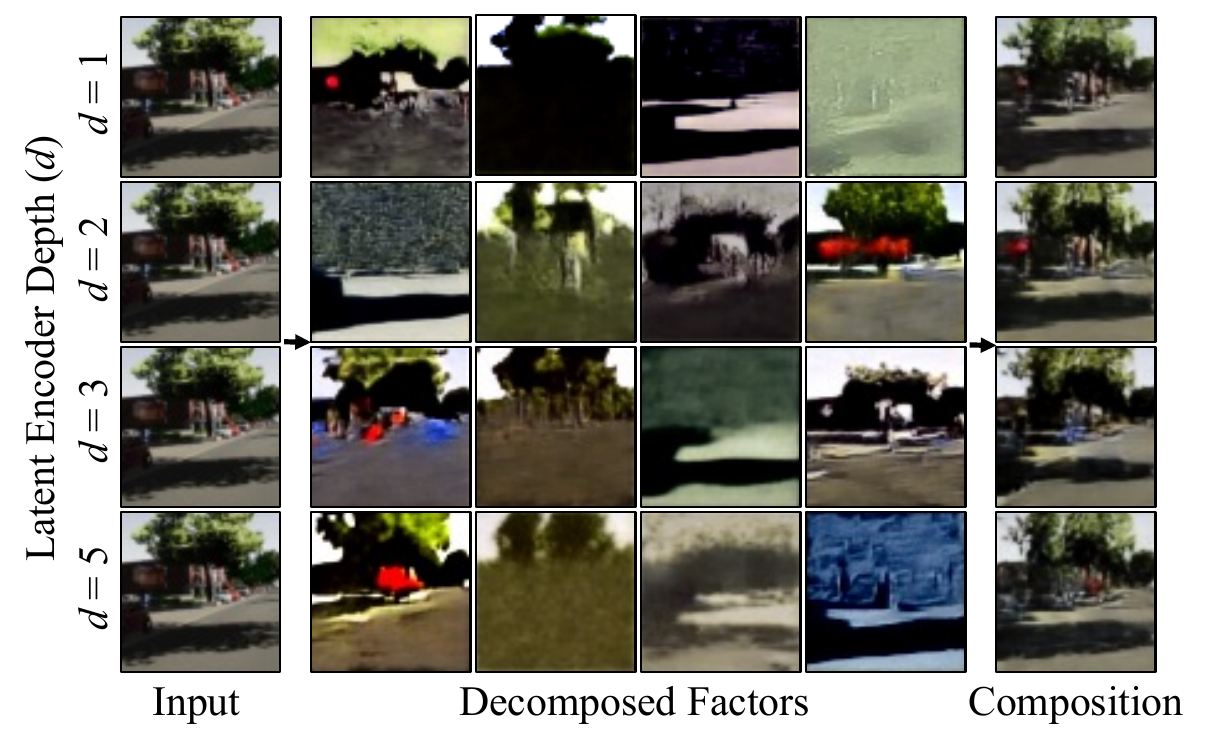}
\end{center}
\vspace{-15pt}
\caption{\textbf{Impact of latent encoder depth on VKITTI.} Encoders with different depths, denoted as $d$, can learn similar decomposed factors, including shadows, background, etc.}
\vspace{-15pt}
\label{fig:encoder_depth}
\end{figure}

\noindent\textbf{Impact of Latent Encoder Depth}
To see how the latent encoder design impacts decomposition performance, we tested decomposition on VKITTI using different encoder depths. Specifically, we experimented with an encoder of depth 1, $\ie$, 1 residual block and convolution layer, as well as depth 2, depth 3 (the default value we used in the main paper), and depth 5, with results shown in Figure \ref{fig:encoder_depth}. We demonstrate that our method is not sensitive to encoder depth changes, as the encoders with different depths learn similar decomposed factors, including shadows, backgrounds, etc.

\section{Model Details}
\label{sup:model_details}

We used the standard U-Net architecture from \cite{ho2020denoising} as our diffusion model. To condition on each inferred latent $\vz_k$, we concatenate the time embedding with encoded latent $\vz_k$, and use that as our input conditioning. In our implementation, we use the same embedding dimension for both time embedding and latent representations. Specifically, we use $256$, $256$, and $16$ as the embedding dimension for both timesteps and latent representations for  CelebA-HQ, Virtual KITTI $2$, and Falcor3D, respectively. For datasets CLEVR, CLEVR Toy, and Tetris, we use an embedding dimension of $64$.

To infer latents, we use a ResNet encoder with hidden dimension of $64$ for Falcor3D, CelebA-HQ,  Virtual KITTI $2$, and Tetris, and hidden dimension of $128$ for CLEVR and CLEVR Toy. In the encoder, we first process images using $3$ ResNet Blocks with kernel size $3\times3$. We downsample images between each ResBlock and double the channel dimension. Finally, we flatten the processed residual features and map them to latent vectors of a desired embedding dimension through a linear layer.

\section{Experiment Details}
\label{sup:experiment}

In this section, we first provide dataset details in \sect{subsec:dataset}. We then describe training details for our baseline methods in \sect{subsec:baselines}. Finally, we present training and inference details of our method in \sect{subsec:training} and \sect{subsec:inference}.

\subsection{Dataset Details}
\label{subsec:dataset}

\begin{table}[t!]
    \centering
    \small
    \setlength{\tabcolsep}{3mm}
    \label{table:ablation}
    \begin{tabular}{lc}
    \toprule  
    Dataset  & Size  \\
    \midrule
    CLEVR & 10K  \\  
    CLEVR Toy & 10K  \\  
    CelebA-HQ & 30K \\  
    Anime & 30K \\
    Tetris & 10K \\
    Falcor3D & 233K \\
    KITTI & 8K \\
    Virtual KITTI 2 & 21K \\
    \bottomrule
    \end{tabular}
    \vspace{-5pt}
    \caption{Training dataset sizes.}
    \label{tab:train_size}
\end{table}

Our training approach varies depending on the dataset used. Specifically, we utilize a resolution of $32\times32$ for Tetris images, while for other datasets, we use $64\times64$ images. The size of our training dataset is presented in Table \ref{tab:train_size} and typically includes all available images unless specified otherwise.

\begin{table}[h]
    \centering
    \small
    \setlength{\tabcolsep}{2mm}
    \begin{tabular}{lcccc}

    \toprule
      {\bf \multirow{2}{*}{\cm{Model}}} &  \multicolumn{2}{c}{\cm{\bf CLEVR}} &
      \multicolumn{2}{c}{\cm{\bf CLEVR Toy}} \\
      \cmidrule(lr){2-3} \cmidrule(lr){4-5} 
      
      & \cm{FID $\downarrow$} & \cm{KID $\downarrow$} & \cm{FID $\downarrow$} & \cm{KID $\downarrow$} \\
     \midrule
     \cm{COMET} & \cm{$98.27$} & \cm{$0.110$} & \cm{$192.02$} & \cm{$0.250$}  \\
     \cm{Ours} & \cm{$\bf75.16$} & \cm{$\bf0.086$} & \cm{$\bf52.03$} & \cm{$\bf0.052$}  \\
    \bottomrule
    \end{tabular}
    \vspace{5pt}
    
    \captionof{table}{\small \cm{ \textbf{Cross-dataset quantitative metrics.} For evaluating cross-dataset recombination (CLEVR combined with CLEVR Toy), because there is no ground truth for recombined images, we computed FID and KID scores of generated images against the original CLEVR dataset and CLEVR Toy dataset. Our approach achieves better scores for both datasets compared to COMET, which suggests that our generations are more successful in recombining objects from the original datasets.
}
}
    \label{tab:cross}
    \vspace{-10pt}
\end{table}

\textbf{Anime.}~\citep{danbooru2019Portraits} When creating the multi-modal faces dataset, we combined a $30,000$ cropped Anime face images with $30,000$ CelebA-HQ images.

\textbf{Tetris.}~\citep{greff2019multi} We used a smaller subset of 10K images in training, due to the simplicity of the dataset.

\textbf{KITTI.}~\citep{Geiger2012CVPR} We used $8,008$ images from a scenario in the the Stereo Evaluation 2012 benchmark in our training.

\textbf{Virtual KITTI $2$.}~\citep{cabon2020virtual}
We used $21,260$ images from a setting in different camera positions and weather conditions.

\subsection{Baselines} 
\label{subsec:baselines}

\textbf{Info-GAN~\citep{Chen2016InfoGAN}.} We train Info-GAN using the default training settings from the official codebase at \href{https://github.com/openai/InfoGAN}{https://github.com/openai/InfoGAN}.

\textbf{$\beta$-VAE~\citep{Higgins2017Beta}.} We utilize an unofficial codebase to train $\beta$-VAE on all datasets til the model converges. We use $\beta = 4$ and $64$ for the dimension of latent $\vz$. We use the codebase 
in \href{https://github.com/1Konny/Beta-VAE}{https://github.com/1Konny/Beta-VAE}.

\textbf{MONet~\citep{burgess2019monet}.} We use an existing codebase to train MONet models on all datasets until models converge, where we specifically use $4$ slots, and $64$ for the dimension of latent $\vz$. We use the codebase in \href{https://github.com/baudm/MONet-pytorch}{https://github.com/baudm/MONet-pytorch}.

\textbf{COMET~\citep{du2021comet}.} We use the official codebase to train COMET models on various datasets, with a default setting that utilizes $64$ as the dimension for the latent variable $\vz$. Each model is trained until convergence over a period of $100,000$ iterations. We use the codebase in \href{https://github.com/yilundu/comet}{https://github.com/yilundu/comet}.

\textbf{Slot Attention~\citep{locatello2020objectcentric}.} We use an existing PyTorch implementation to train Slot Attention from \href{https://github.com/evelinehong/slot-attention-pytorch}{https://github.com/evelinehong/slot-attention-pytorch
}.

\textbf{GENESIS-V2~\citep{engelcke2021genesisv2}.} We train GENESIS-V2 using the default training settings from the official codebase at \href{https://github.com/applied-ai-lab/genesis}{https://github.com/applied-ai-lab/genesis
}.

\subsection{Training Details}
\label{subsec:training}

We used standard denoising training to train our denoising networks, with $1000$ diffusion steps and squared cosine beta schedule. In our implementation, the denoising network $\epsilon_\theta$ is trained to
directly predict the original image $\vx_0$, since we show this leads to better performance due to the similarity between our training objective and autoencoder training.

To train our diffusion model that conditions on inferred latents $\vz_k$, we first utilize the latent encoder to encode input images into features that are further split into a set of latent representations $\{\vz_1, \hdots, \vz_K\}$. For each input image, we then train our model conditioned on each decomposed latent factor $\vz_k$ using standard denoising loss.

\cm{Regarding computational cost, our method uses $K$ diffusion models, so the computational cost is $K$ times that of a normal diffusion model. In practice, the method is implemented as $1$ denoising network that conditions on $K$ latents, as opposed to $K$ individual denoising networks. One could significantly reduce computational cost by fixing the earlier part of the network, since latents would only be conditioned on in the second half of the network. This would likely achieve similar results with reduced computation. In principle, we could also parallelize $K$ forward passes to compute $K$ score functions to reduce both training and inference time.}

Each model is trained for $24$ hours on an NVIDIA V100 32GB machine or an NVIDIA GeForce RTX 2080 24GB machine. We use a batch size of $32$ when training.

\subsection{Inference Details}
\label{subsec:inference}

When generating images, we use DDIM with 50 steps for faster image generation.

\textbf{Decomposition.}
To decompose an image $\vx$, we first pass it into the latent encoder $\text{Enc}_{\theta}$ to extract out latents $\{\vz_1, \cdots, \vz_K\}$. For each latent $\vz_k$, we generate an image corresponding to that component by running the image generation algorithm on $\vz_k$. 

\textbf{Reconstruction.}
To reconstruct an image $\vx$ given latents $\{\vz_1, \cdots, \vz_K\}$, in the denoising process, we predict $\epsilon$ by averaging the model outputs conditioned on each individual $\vz_k$. The final result is a denoised image which incorporates all inferred components, \ie, reconstructs the image.

\textbf{Recombination.} To recombine images $\vx$ and $\vx'$, we recombine their latents $\{\vz_1, \cdots, \vz_K\}$ and $\{\vz_1', \cdots, \vz_K'\}$. We select the desired latents from each image and condition on them in the image generation process, \ie, predict $\epsilon$ in the denoising process by averaging the model outputs conditioned on each individual latent.  

To additively combine images $\vx$ and $\vx'$ so that the result has all components from both images, $\eg$, combining two images with $4$ objects to generate an image with 8 objects, we modify the generation procedure. In the denoising process, we assign the predicted $\epsilon$ to be the average over all $2\times K$ model outputs conditioned on individual latents in $\{\vz_1, \cdots, \vz_K\}$ and $\{\vz_1', \cdots, \vz_K'\}$. This results in an image with all components from both input images.

\onecolumn

\end{document}